PaddleMaterials: An End-to-End Deep Learning Toolkit for Materials Science

Xuwei Liu[1*], Zhimin Zhang[1], Xiaoguang Hu[1], Zhongwei Liu[2], Yuanhui Sun[2], Jun Huo[1], Hua Wu[1], Haifeng Wang[1*]
[1] Baidu Inc., Beijing, 100094, China
[2] Suzhou National Laboratory, Suzhou, 215123, China
[*]Corresponding authors: liuxuwei01@baidu.com, wanghaifeng@baidu.com

# Abstract

Machine learning tasks in materials science support structure design, atomistic simulation and experimental characterization, but the associated models, data representations and research workflows are usually developed independently. Integrating these tasks requires not only unifying software interfaces but also preserving the specific features of each task and making the scientific semantics and computational behavior of cross-task combinations verifiable. To this end, we present PaddleMaterials, an open-source unified framework for model development and scientific application. At its core, the framework explicitly models the computational structure and scientific semantics of tasks and brings model composition and execution mechanisms into a single system, thereby improving the verifiability of cross-task operations and characterizing performance under a unified execution system. Through shared mechanisms for data processing, training and evaluation, pretrained-model management and task application, the platform organizes six task classes: property prediction, structure generation, machine-learning interatomic potentials (MLIPs), electronic-structure fields, spectrum elucidation and microscopy image processing. Each task retains its own structure, graph, field, spectrum or image representation, while scientific-role records preserve the physical quantities, units, aggregation conventions and reference information at its input and output boundaries. The platform uses deep-learning compiler technology to accelerate model execution: through graph optimization, operator fusion and code generation for the target hardware, a model can switch from immediate dynamic-graph execution to compiled execution without changes to its definition. We validated both execution paths on 65 model instances in 11 model groups: relative to dynamic-graph execution, every model instance in 8 of the groups ran faster under compiled execution, with group median speedups of 11%–224% and group maximum speedups of 11%–336%. In two case studies, the inorganic material $LiMnO_2$ and the organic molecule toluene, the platform linked structure generation or spectrum elucidation, MLIP relaxation, and property and electron-density prediction into continuous workflows, with same-geometry density functional theory (DFT) calculations as external references. For $LiMnO_2$, the formation-energy predictions on the generated and relaxed structures differed from the same-geometry PBE+U results by no more than 7 meV atom$^{-1}$, and the total computational cost of the model steps was less than 1/20 of that of one DFT single-point calculation. These results show that, compared with existing platforms that mainly target structure–property prediction, interatomic potentials or structure generation, PaddleMaterials also brings electron-density, spectral and microscopy-image tasks into one framework while keeping their dedicated representations, makes cross-task combinations verifiable through explicit scientific-semantics conventions, and lets heterogeneous models share one compiled execution backend by changing a single execution setting, which keeps the barrier to use low and raises execution efficiency.

# Introduction

Discovering new materials with target properties and establishing predictive links from composition, structure, processing history and service conditions to material properties (that is, structure–property relationships) are among the central goals of materials science research. Establishing such relationships often relies on multiscale simulation and can be cast as a high-dimensional coupled problem defined jointly by material parameters, cross-scale state variables and physical constraints. However, the solution space is enormous: it involves continuous high-dimensional state and parameter spaces and is accompanied by a combinatorial explosion arising from composition, structure, processing and other variables. As a result, systematic exploration that relies on exhaustive sampling or high-fidelity simulation is generally limited by the curse of dimensionality and high computational cost. Experiments and first-principles calculations can provide references, but they can hardly cover such a vast design space directly.

Deep learning offers a new route out of this dilemma. With its capabilities in high-dimensional function approximation and representation learning, it can combine regularities in data with physical priors to achieve efficient search, compact representation and generalizable inference in complex spaces. Materials science enabled by artificial intelligence (AI for Materials) has gradually become an important research approach that connects materials data, computational models and design goals. Turning these methods into reusable research capabilities, however, first requires coping with the diversity of model architectures. For materials modeling, graph neural networks and their equivariant architectures encode interatomic relationships and geometric symmetries to learn material properties and interactions, and diffusion models learn structural distributions and conditional dependencies to generate candidate structures [1, 2]. For scientific computing, physics-informed neural networks (PINNs) incorporate governing equations and initial and boundary conditions into the training objective to solve forward and inverse problems of partial differential equations, and Fourier neural operators (FNOs) learn mappings between function spaces to provide reusable approximate solvers [3, 4]. This diversity continues to extend to frontier deep learning architectures: flow matching generates crystal and molecular structures through more stable continuous transformations [5, 6]; large language model architectures encode materials as sequences to unify generation and property inference [7, 8]; multimodal models fuse structural, textual, spectral and other information to build cross-modal representations [9]; and foundation models pretrained across material systems provide transferable approximations for universal machine-learning interatomic potentials [10, 11]. These methods stem from the transfer of general-purpose architectures to scientific problems and also rely on dedicated designs oriented toward physical laws. Model development therefore requires coordinated trade-offs among network architecture, data representation, physical inductive bias, training objective and inference procedure, and it also requires the host platform to keep accommodating continually evolving frontier architectures.

Model diversity is further intertwined with the heterogeneity of material objects, research scales and experimental observation modes, giving AI for Materials a pronounced multitask character (Fig. 1). Research tasks have expanded from conventional structure–property prediction to multiple directions, including structure generation, spectrum elucidation, atomic-resolution and microscopy image analysis, electron-density prediction, machine-learning interatomic potentials, retrosynthesis analysis, and cross-scale microstructure and process optimization. Different tasks deal with different scientific objects and target variables, and their data must be expressed in multiple modalities, including graphs that describe interatomic relationships (molecular graphs and periodic crystal structures), sequence representations (structural encodings such as SMILES/SELFIES and composition formulas), continuous spatial fields (such as electron density), spectra, microscopy and atomic-resolution images, and even text that carries synthesis and literature knowledge [12–14]. It is therefore difficult to rely on a single general-purpose representation or a single modeling paradigm. Meanwhile, these tasks often rely on different types of learning models and inference mechanisms, and they can be combined and chained in specific materials research and development workflows. Consequently, an AI for Materials platform needs not only to cover diverse materials tasks but also to support both model development (data processing, construction of scientific representations, and model training and evaluation) and model application (applying existing models to new material structures, simulation conditions and experimental characterization) [1, 2, 12].

Existing open-source platforms and benchmarks have advanced this goal from different angles (Table 1), but three limitations remain. First, functional coverage is limited. MatGL organizes property models and interatomic potentials around graph deep learning [14]; M²Hub provides unified benchmarks for screening, inverse design and simulation [13]; Matbench establishes comparable benchmarks for materials property prediction through standardized datasets and evaluation tasks [15]; and the Open MatSci ML Toolkit supports equivariant model development and scalable training centered on Open Catalyst [16]. Each covers part of the task spectrum, and tasks such as electron-density prediction, spectrum elucidation and microscopy image processing are not yet included, so researchers often have to assemble several tools. Second, the engineering loop is not closed. Even when the required models have been implemented, some tools remain centered on specific methods or on benchmark construction, leaving gaps between model development and model application; training control, model saving and invocation, and inference deployment often require further development by users. Third, the underlying architectural abstractions and usability are insufficient. The key to supporting diverse AI for Materials tasks is to share the software control flow while isolating and preserving non-negligible scientific differences such as coordinates, units, physical conservation laws and boundary conditions. Existing tools lack clear underlying architectural abstractions, so general engineering code is coupled with task-specific physical logic: general-purpose platforms struggle to express these differences, whereas platforms built for a single task class struggle to reuse their software mechanisms for new tasks. The ecosystem also lacks comprehensive examples that run from scientific data loading and model construction to directly calling pretrained models to solve practical problems, and the shortage of systematic documentation and examples makes the tools hard to learn and limits their adoption among materials researchers. The field therefore needs an end-to-end AI infrastructure that unifies multimodal representations and multitask workflows, provides a complete engineering chain from model development to application, and offers a rich set of pretrained models and ready-to-use examples.

PaddleMaterials is positioned as an end-to-end toolkit for AI for Materials to meet this need. Built on PaddlePaddle, it targets research objects such as inorganic crystals and organic molecules, organizes property prediction, structure generation, machine-learning interatomic potentials, electron-density prediction, spectrum elucidation and microscopy image enhancement under a unified entry point, and provides the accompanying datasets, model implementations and pretrained models. Its main design principles are modularity, registry-based extensibility, single-file model organization, standardized execution workflows and interoperability with scientific computing tools such as pymatgen, RDKit and ASE. Software mechanisms such as data loading, training control, and model saving and invocation are reused across tasks as shared infrastructure, whereas coordinates, units, boundary conditions, prediction targets and evaluation methods are specified explicitly by each task, and the domain-specific representations, predictors and samplers that each task requires are retained. This establishes a reusable engineering separation between shared workflows and task differences, and comprehensive cross-task examples lower the barrier to use. PaddleMaterials thus integrates multitask coverage, an end-to-end chain from model development to application, reusable software mechanisms and task-level scientific conventions into one low-barrier toolkit, rather than adding another single model or benchmark.

This paper first gives a unified formal description of materials tasks, reviews the heterogeneous tasks covered by the platform and the requirements they place on a unified software design, and introduces the platform architecture and core software abstractions designed to meet these requirements. It then reports comparisons with reference implementations on public benchmarks and experiments on compiled execution based on deep-learning compiler technology, and finally examines the complete workflow from material objects to model execution through cross-task application cases. In the evaluation, we discuss functional coverage, the correctness and efficiency of software execution, consistency with reference results, and utility for materials research and development separately, so as not to conflate the availability of software functions, numerical consistency, runtime performance and practical scientific value into conclusions at a single level.

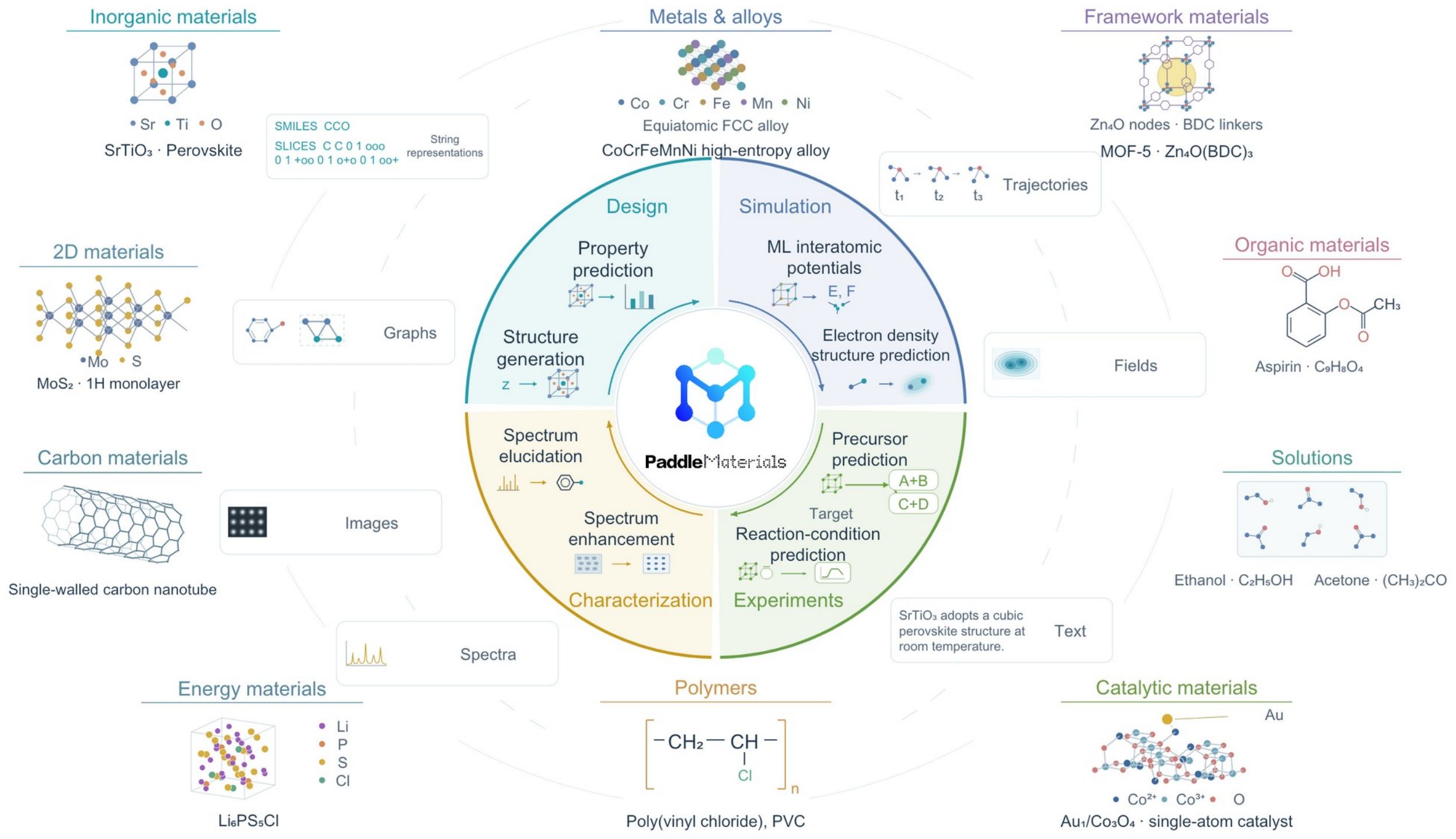

Fig. 1 | PaddleMaterials: an AI framework for diverse material classes, data representations and materials research tasks. From the outer to the inner ring, the figure presents representative material classes, data representations and AI tasks for materials research. Material classes include inorganic materials, metals and alloys, framework materials, organic materials, solutions, catalytic materials, polymers and energy materials. Data representations comprise strings, graphs, fields, images, spectra, trajectories and text. Tasks span materials design, simulation, characterization and experimentation, including property prediction, structure generation, machine-learning interatomic potentials, electron-density prediction, spectrum elucidation and enhancement, precursor prediction and reaction-condition prediction. The examples illustrate the scope of materials research and do not imply one-to-one correspondences between the rings.

Table 1 | Scientific scope and reported evaluation of selected materials machine learning platforms [13, 14, 16].

| Platform | Scientific scope | Shared infrastructure | Evaluation |
|---|---|---|---|
| M²Hub | Virtual screening, inverse design and molecular simulation, formulated as property prediction, generative materials design and machine-learning force-field tasks. | Curated datasets and graph preprocessing; common training and evaluation pipelines; split protocols suited to each dataset; generative metrics and oracle functions. | Representative property prediction benchmarks under applicable random, composition and system splits; training runtime comparisons. |
| Open MatSci ML Toolkit | Graph and point-cloud learning for energies, forces and other material properties across multiple datasets; CDVAE crystal generation supported in version 1.0. | Shared data, model and task abstractions; graph and point-cloud transforms; common training and evaluation; distributed execution; multi-task and multi-dataset learning. | OC20 validation/test errors and compute scaling; single-task, multi-task and multi-dataset benchmarks; CDVAE generation metrics in the v1.0 release report. |
| MatGL | Graph deep learning for molecular and crystalline property prediction and interatomic potentials, with interfaces for atomistic simulation. | Structure-to-graph pipelines; reusable invariant/equivariant GNN layers; PyTorch Lightning-based training; pretrained models; ASE and LAMMPS interfaces. | Property and potential benchmarks; transfer learning and extrapolation tests; atomistic application validation; inference timing and scaling of molecular dynamics simulations with system size. |
| PaddleMaterials (this study) | Model development and application for materials such as inorganic crystals and organic molecules, covering six task classes: property prediction, structure generation, machine-learning interatomic potentials, electron-density prediction, spectrum elucidation and microscopy image enhancement. | Shared mechanisms for data processing, training/evaluation and task application; pretrained-model packages; explicitly documented scientific input/output conventions; optional CINN compiled execution for selected model computations. | Comparison of 11 migrated models with the results of the original implementations on public benchmarks for six task classes; two cross-task application case studies, $LiMnO_2$ and toluene, with same-geometry DFT references; eager–CINN dual-path verification and performance tests of 65 model instances. |

# Results

## Unified formal description of tasks

We formalize the task organization of PaddleMaterials at two levels: task-specific definitions and shared platform capabilities. PaddleMaterials currently covers machine learning tasks in materials science (hereafter, materials tasks) such as property prediction, structure generation, machine-learning interatomic potentials, electron-density prediction, spectrum elucidation and microscopy image enhancement, which are markedly heterogeneous in their input objects, output forms, model architectures and scientific goals. Let the set of materials tasks supported by the platform be $T=\{\tau_1,\tau_2,\ldots,\tau_N\}$. For any specific task $\tau\in T$, we define

$$\tau=(X_\tau,Y_\tau,f_{\theta\tau},L_\tau,M_\tau,S_\tau)$$

where $X_\tau$ denotes the input space of the task, $Y_\tau$ the output space, $f_{\theta_\tau}$ the task model, $L_\tau$ the training objective and $M_\tau$ the evaluation method, and $S_\tau$ denotes the scientific-semantics convention of the task inputs and outputs, which includes the scientific objects or physical quantities represented as well as the applicable units, reference baselines, coordinate representations and measurement bases. Each materials task specifies these elements separately, retaining its own data representation, model, learning objective and scientific meaning.

As a counterpart to these task-specific meanings, the platform provides a set of shared capabilities, denoted

$$\Phi=(\Phi_{data},\Phi_{model},\Phi_{train},\Phi_{eval},\Phi_{use})$$

where $\Phi_{data}$ provides data construction, representation and batching; $\Phi_{model}$ provides model registration, construction and resource loading; $\Phi_{train}$ provides organization and optimization during training (training-workflow orchestration, optimizers, learning-rate scheduling and loss computation); $\Phi_{eval}$ computes metrics according to the evaluation method specified by the task; and $\Phi_{use}$ provides model inference, which can be further expressed as the set of operations $\Phi_{use}=\{\Phi_{pred},\ \Phi_{samp},\Phi_{\mathrm{sim}}\}$, corresponding to prediction, generative sampling and numerical solution using predictive models, respectively. Each task reuses the corresponding operation workflows and specifies its task-specific components and execution parameters through the task configuration.

The shared training workflow organizes the task-specific data, model and training objective, and its parameter optimization process is expressed as

$$\hat{\theta}_\tau=\Phi_{train}(D_\tau,\ f_{\theta_\tau},L_\tau;c_\tau^{train})$$

where $D_\tau$ is the training data and $c_\tau^{train}$ is the training configuration, which contains the optimization method, training parameters and randomness settings. For task evaluation, $\Phi_{eval}$ computes the corresponding metrics according to $M_\tau$. In the existing training entry point, the data loader, model, optimizer and metric components are built from the configuration, and training, validation and testing are organized by Trainer.

The model inference stage employs the parameters $\hat{\theta}_\tau$, obtained from training or loaded from pretrained resources. For prediction and generation tasks with given inputs, the execution processes are, respectively:

$$\hat{y}_\tau=\Phi_{pred}(f_{\hat{\theta}_\tau},x;c_\tau^{pred}),\ \ x\in X_\tau,\ \ \hat{y}_\tau\in Y_\tau$$

$$\tilde{y}_\tau=\Phi_{samp}(f_{\hat{\theta}_\tau},h,\ \xi;c_\tau^{samp}),\ \ \tilde{y}_\tau\in Y_\tau$$

where $\hat{y}_\tau$ is the prediction for a given input $x$ and $\tilde{y}_\tau$ is a generated candidate; $h$ is the conditioning input of the generation task (including composition constraints, target properties or observed signals), which is set to the empty condition for unconditional generation; $\xi$ represents the randomness in the sampling process; and $c_\tau^{pred}$ and $c_\tau^{samp}$ are the prediction and sampling configurations, respectively. Each of the two operations organizes input processing, model invocation and result conversion, while retaining task-specific execution procedures. For example, the property prediction entry point processes structural inputs through PropertyPredictor, and the structure generation entry point samples through StructureSampler conditioned on formulas, atom counts or properties.

To meet the computational need in scientific machine learning (SciML) for model predictions to take part in numerical solution, we abstract this mode of model use as $\Phi_{\text{sim}}$, which links model computation with numerical algorithms through a prediction interface. Let $\Phi_{pred}^{\tau}$ denote the prediction interface that binds the task model and prediction configuration and satisfies the input/output conventions of the corresponding solver. The computational result of this class of methods and its associated scientific semantics are expressed as

$$r_\tau^{\text{sim}} = \left( \Phi_{\text{sim}} \left( x_\tau^{\text{sim}} ; \Phi_{pred}^{\tau}, c_\tau^{\text{sim}} \right), \sigma_\tau^{\text{sim}} \right)$$

where $x_\tau^{\text{sim}}$ is the scientific computing input, including the applicable initial or boundary conditions, geometric information and physical parameters, and $c_\tau^{\text{sim}}$ specifies the numerical problem, the solution algorithm and the scientific computing configuration. Following the numerical algorithm, $\Phi_{\text{sim}}$ organizes the computation and obtains the required predicted quantities through $\Phi_{pred}^{\tau}$ depending on the algorithm and the current solution state. The term $\sigma_\tau^{\text{sim}}$ denotes the scientific semantics of the result, determined jointly by the semantic conventions of the predicted quantities, the definition of the numerical problem and the output conventions.

## Coverage of heterogeneous materials-science tasks and representative implementations

Within this formal framework, PaddleMaterials integrates representative implementations of six task classes, namely property prediction, structure generation, machine-learning interatomic potentials, electron-density prediction, spectrum elucidation and microscopy image enhancement, and provides the corresponding entry points for data processing, training, evaluation and application. These tasks differ in research objects, inputs and outputs, and learning objectives, yet they need to share data processing, training and application workflows while retaining task-specific data representations, model architectures, learning objectives and evaluation methods. In computational terms, these tasks involve different types of mapping and distribution modeling. Property prediction maps a materials representation to a target property, $y = f_\theta(X)$. Machine-learning interatomic potentials predict the energy and forces of an atomic configuration; for energy-based models, the atomic forces are obtained as the negative gradient of the potential energy, $F_i = -\dfrac{\partial E_\theta(Z, R, L)}{\partial r_i}$. Electronic-structure tasks predict the electron density, the Hamiltonian or other electronic-structure representations; electron-density prediction can be written as $\rho(r) = f_\theta(X, r)$. Spectrum enhancement maps degraded signals or images to restored observations, $Y_{\text{degraded}} \rightarrow Y_{\text{enhanced}}$. Structure generation models unconditional or conditional structure distributions and samples candidates in the structure space. Spectrum elucidation infers candidate structures from observations, $X \sim p_\theta(X / S)$. Supplementary Table S1 summarizes the general scope of these tasks and lists their representative models and datasets in PaddleMaterials.

From the perspective of materials-science applications, these task families correspond to the following problems: predicting formation energies, band gaps, elastic moduli and molecular quantum-chemical properties from crystal or molecular structures; generating candidate structures from chemical compositions or target properties to supply candidates for materials screening and inverse design; learning physical quantities such as the energy, forces and stress of atomic configurations for structural relaxation and molecular dynamics; predicting electronic-structure information such as electron densities, Hamiltonians or energy spectra to support the analysis of electronic properties; inferring candidate molecular or materials structures from spectroscopic information such as nuclear magnetic resonance spectra; and denoising and reconstructing spectral signals or microscopy images to support spectral peak identification, atom localization and defect analysis. Each task retains its own definitions of physical quantities, structural constraints and evaluation methods, and its implementation is not restricted to a single model or experimental modality.

Tasks differ not only in data formats or output layers but also in scientific representations, learning objectives and computational workflows. For example, crystal generation must handle the lattice and atomic coordinates, energy-based interatomic potentials obtain atomic forces from derivatives of the potential energy, electron-density prediction involves spatial-field representations, spectrum elucidation must deal with the multiplicity of solutions in structure inference, and microscopy image restoration must preserve the spatial information relevant to subsequent analysis. A unified platform therefore has to share software workflows while preserving these task-specific scientific semantics. Model architectures need not be bound to fixed tasks: the same task can use models with different inductive biases, and the same model architecture can be adapted to the output requirements and learning objectives of different tasks while retaining its core representation and inductive biases. For example, SphereNet$^{paddle}$ can be used both for QM9 molecular property prediction and for MD17 energy–force modeling.

These differences place five concrete requirements on a unified platform. First, input representations differ, so the platform must unify the construction from raw scientific data to task representations while keeping each task's own representation. Second, computational structures and training objectives differ (for example, atomic forces are obtained by differentiating the potential energy with respect to the coordinates, field values are queried at spatial coordinates, and generative models sample iteratively), so the platform must let each model define its own computation and fix only the interfaces through which it joins the shared workflows. Third, the modes of use differ, including training, prediction, generation and coupling with external simulation programs, so the platform must organize execution for each mode. Fourth, outputs must state the physical quantities, units, aggregation levels and reference conventions they represent, and reused models must carry data-processing states such as normalization statistics and vocabularies. Fifth, the computational forms of the tasks vary widely and are often invoked repeatedly in simulation and generation, so execution efficiency should be optimized uniformly without changing model definitions.

## Architecture of the PaddleMaterials platform

In response to these requirements, PaddleMaterials is divided by software responsibility into three interconnected layers, namely the task layer, the orchestration layer and the components layer, all built on the PaddlePaddle deep learning framework (Fig. 2). The task layer launches and routes specific tasks, the orchestration layer organizes training, prediction, generation or simulation processes, and the components layer provides concrete implementations of models, data, optimization, evaluation and other functions, together with their configuration-driven construction mechanisms. Researchers describe the model, data, optimization and evaluation processes through configurations; the task layer serves as the entry point that launches tasks accordingly, and the orchestration layer instantiates the components and organizes execution processes such as training, prediction, sampling or atomistic simulation. This design separates task entry points, the execution lifecycle and functional implementations from one another, so that different materials tasks can share software infrastructure while retaining their differences in model architecture and scientific representation.

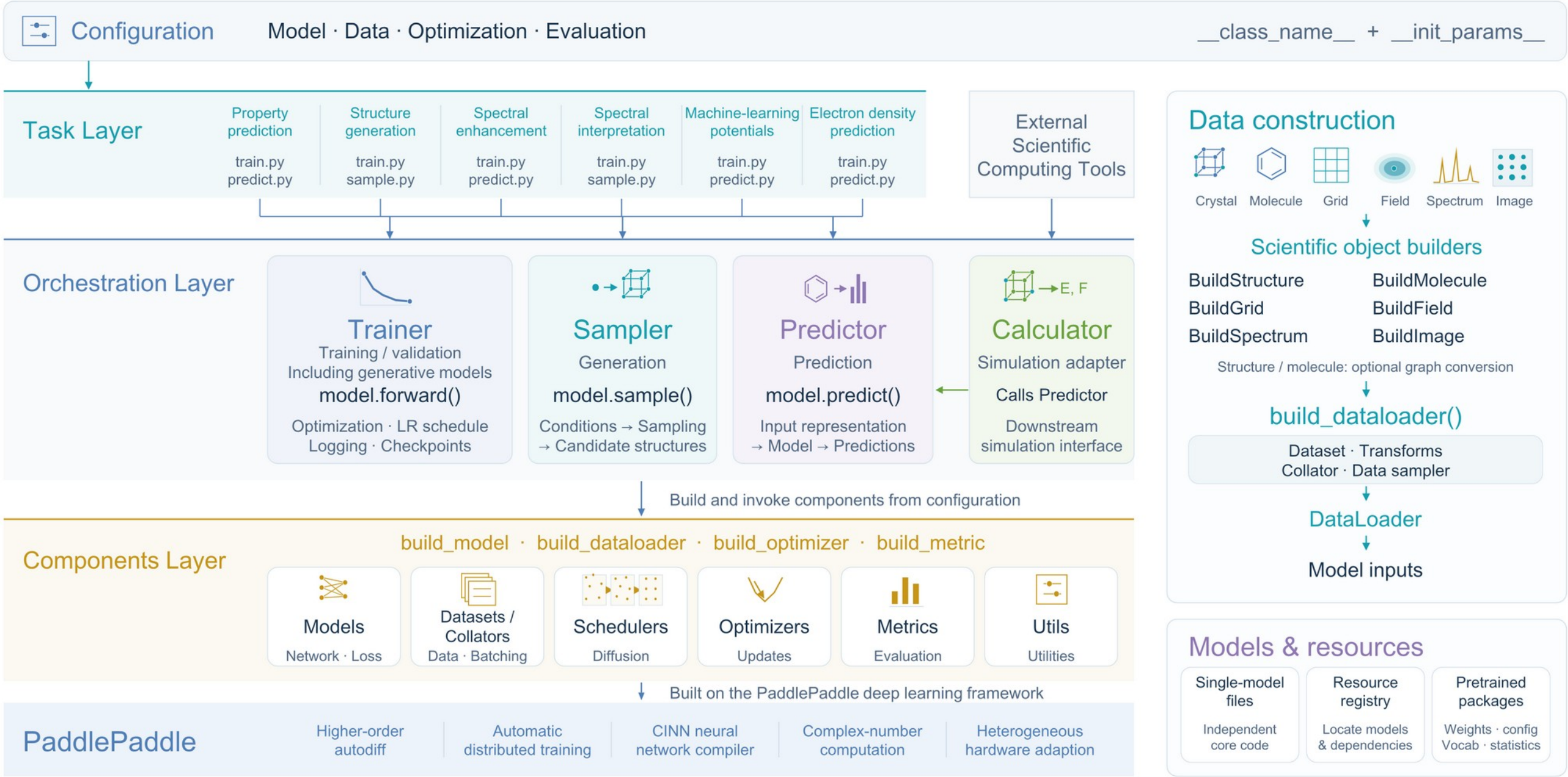

Fig. 2 | Architecture and unified workflow of PaddleMaterials. PaddleMaterials adopts a layered architecture comprising the task layer, the orchestration layer and the components layer, built on a PaddlePaddle backend. Configuration-driven task entry points support training, prediction and sampling through dedicated runtime abstractions, including Trainer, Predictor, Sampler and Calculator. Reusable components provide model construction, data processing, optimization, scheduling, evaluation and utilities. Scientific data builders accommodate heterogeneous material representations, while model and resource management supports extensibility and pretrained-model reuse. The PaddlePaddle backend provides tensor computation, automatic differentiation, distributed execution and CINN-based compilation.

The task layer consists of task scripts such as train.py, predict.py and sample.py, which mainly read the configuration, initialize the corresponding runtime components and launch specific tasks, while keeping network architectures and task-specific scientific modeling logic out of the scripts as far as possible. The main role of the task layer is to translate the experiment description given by the user into calls to the underlying runtime.

To serve the different modes of use, the orchestration layer organizes the execution of models across different lifecycle stages: Trainer handles training and validation, whereas Predictor, Sampler and Calculator host the application execution modes of prediction, generation and simulation, respectively. Trainer organizes common procedures such as the training and validation loops, optimizers and learning-rate scheduling, logging, and checkpoint saving and restoration. Trainer does not depend on the internal network architecture of a model but interacts with the model through an agreed interface: the model component receives batched data and returns the loss terms and predictions required for training, where the total loss is used for automatic differentiation and parameter updates and the predictions are passed on to evaluation and subsequent analysis. Generative models likewise connect to Trainer through forward() during training and thereby reuse the unified training and validation workflow. Trainer is thus responsible for the execution lifecycle, whereas the model itself still independently defines its network architecture, physical inductive biases and task-related training objectives. In the model application stage, Predictor, Sampler and Calculator serve different types of scientific tasks. Predictor serves predictive tasks: it converts new material objects or task inputs into the representations required by the model and organizes model invocation and the return of results through the predict() interface provided by the model. Sampler serves generation tasks: through the sample() interface provided by the model, it organizes condition construction, iterative sampling and the output of candidate structures, independently of the Trainer path used in the training stage. Calculator acts as an adapter between surrogate models and downstream scientific computing and simulation workflows; through Predictor, it turns the predictive capability of a model into an interface that downstream scientific computing programs can call, thereby supporting reuse across a wide range of materials research workflows.

The components layer provides concrete implementations of models, datasets and batching components, schedulers, optimizers, evaluation metrics and general utilities (Models, Datasets/Collators, Schedulers, Optimizers, Metrics and Utils) and is responsible for building runtime instances from configurations. Builder functions such as build_model, build_dataloader, build_optimizer and build_metric instantiate the corresponding components from the experiment configuration, separating decisions on model selection, data organization, optimization strategy, evaluation method and the like from the runtime control logic. With the object construction mechanism based on __class_name__ and __init_params__, researchers can replace or combine existing components by adjusting the configuration, without modifying the common training backbone. At its base, this layer relies uniformly on PaddlePaddle for tensor computation, automatic differentiation and parallel execution, and it can further use the PaddlePaddle compiler backend to optimize model training or inference.

To handle the differences in input representations, PaddleMaterials manages the data lifecycle in a unified manner through a two-level construction mechanism that operates on scientific objects and on data pipelines. First, a set of builders for domain objects converts heterogeneous raw data carrying scientific semantics (structures, molecules, grids, fields, spectra, images and so on) into data representations that tasks can use. BuildStructure and BuildMolecule construct crystal and molecular objects, respectively, which can be further converted into atomic graphs or periodic adjacency representations by a task-specific graph converter; BuildGrid and BuildField organize the correspondence among spatial grids, coordinates and field values to support spatial-field tasks such as electronic structure; BuildSpectrum organizes peak positions, intensities and other spectroscopic features into spectroscopic representations; and BuildImage converts microscopy and spectral images into the corresponding numerical representations. Then, the unified build_dataloader() instantiates the Dataset, data transforms, Collator and data sampler according to the configuration and assembles them into a DataLoader that the model can use directly. What the platform unifies is the way task representations are constructed from raw data, rather than the task representations themselves, so that different scientific objects can share a data-processing framework while retaining their own physical and representational semantics.

To fix output semantics and support model reuse, PaddleMaterials links model development and application through its configuration system, single-file model organization and resource registration mechanism. Configurations describe and instantiate experimental components, the single-file model strategy keeps the core implementation of each model independent, and the registry locates pretrained models and the resources they depend on. On this basis, pretrained-model packages further organize states such as weights, configurations, vocabularies and data statistics in a unified way, supporting consistent reuse in the training and inference stages.

To accommodate differences in computational structure and training objectives, model components in the components layer carry the task-related parameterized computational relationship, that is, $f_{\theta_\tau}$ in the task definition. The definition of a model must specify its input representation, internal computational structure and output semantics, and state its conditions of applicability and its interface conventions for the training and application stages. The input representation specifies how structures, molecules, spatial coordinates, spectroscopic information or other task conditions are organized into computational objects; the internal computational structure specifies modules such as feature embedding, information interaction, state update and readout, and how they are connected; and the output semantics corresponds to the scientific-semantics convention $S_\tau$ of the task, specifying the physical quantities or scientific objects represented by the network outputs, together with conventions such as units, atom-level or structure-level aggregation, energy references and inverse target transforms. The model architecture and its architectural settings determine a parameterized model family; the trained parameters, together with the data-processing states required for computation (such as normalization statistics and vocabularies), determine a specific model instance, which is what a pretrained-model package organizes.

In property prediction on atomic graphs (for example, MEGNet$^{paddle}$ and iComFormer$^{paddle}$), elemental and geometric information is first organized into nodes, edges and other features required by the model; after being updated by interaction modules, these features yield the target prediction through atom-level or structure-level readout. The specific feature construction, update operators and readout structure are specified by each model implementation. Other tasks need to specify their own computational relationships: energy and force models must state whether forces are obtained by differentiating the energy with respect to the coordinates (as in CHGNet$^{paddle}$ and MatterSim$^{paddle}$) or are given by a separate prediction branch; spatial-field models (such as InfGCN$^{paddle}$) must define the computational relationship between the structure representation and the query coordinates or grid, as well as how field values are recovered; and generative models (such as DiffCSP$^{paddle}$ and MatterGen$^{paddle}$) must define the representation of generative states and conditions, the prediction parameterization of the core network and the way in which this prediction enters the sampling update.

A model implementation declares the required computational modules and specifies the data dependencies between them in the forward pass. Architectural settings exposed by existing implementations can be adjusted through the configuration of build_model(), whereas new computational relationships are integrated as new model implementations. Integration requires only two conventions: in the training stage, forward() returns predictions and task-related losses; in the application stage, the model provides predict() or sample() according to the task type, and energy–force models are additionally connected to simulation programs through Calculator. A model that satisfies these conventions can reuse the common execution workflows of Trainer, Predictor, Sampler and Calculator, while its internal network architecture and task-related computations remain independently defined.

## Benchmarks of materials-science models

The model benchmarks of PaddleMaterials cover six task classes: property prediction, machine-learning interatomic potentials (MLIPs), crystal structure generation, electron-density prediction, microscopy image enhancement and spectrum elucidation. Fig. 3 summarizes the evaluation results of the available models for each task category on the corresponding datasets relative to the original implementations, with full values in Supplementary Tables S2–S7; these models share the task, orchestration and components layers of the platform and can be reproduced, compared and invoked within the same workflow.

The property prediction benchmark (Fig. 3a and Supplementary Table S2) measures model performance by prediction error, using the mean absolute error (MAE) for each scalar target. The benchmark covers two types of data: formation energies, band gaps and elastic moduli from MP2018 (the Materials Project 2018 crystal dataset), and quantum-chemical properties from QM9 (the Quantum Machine 9 organic-molecule dataset) [17]; the MAEs of the elastic moduli are computed on $\log_{10}$(GPa).

The MLIP benchmark (Fig. 3b and Supplementary Table S3) likewise uses the MAE, together with the vector/tensor L2 errors defined in the notes to Supplementary Table S3. The datasets comprise eight organic molecules from MD17 (a molecular dynamics trajectory dataset) [18] (Supplementary Table S3a), the MPtrj 2022.9 test set of CHGNet (Crystal Hamiltonian Graph Neural Network; Supplementary Table S3b) and the MPtrj-random-1k crystal test set of MatterSim (Supplementary Table S3c; MPtrj stands for Materials Project trajectory, a dataset of materials relaxation trajectories); the corresponding prediction targets are the energies and forces of the molecules and the energies, forces, stresses and magnetic moments of the crystals. For SphereNet, the energy MAE ($kcal \cdot mol^{-1}$) and the force MAE ($kcal \cdot mol^{-1} \cdot Å^{-1}$) are used.

The crystal structure generation benchmark (Fig. 3c and Supplementary Table S4) is based on MP20 (the Materials Project 20 crystal dataset) and examines the composition-conditioned structure prediction of DiffCSP (Diffusion-based Crystal Structure Prediction) and the de novo generation of DiffCSP and MatterGen. The former is evaluated by the match rate and normalized RMSD, and the latter by the fraction of stable, unique and novel structures (S.U.N.) together with Stable, Unique, Novel and the relaxation RMSD.

The electron-density prediction benchmark (Fig. 3d and Supplementary Table S5) evaluates three-dimensional density fields with the normalized mean absolute error (NMAE, %) on the organic-molecule datasets QM9_EC and MD17 (six molecules) and the crystal dataset MP_EC cubic. The $\text{InfGCN}^{\text{paddle}}$ model retrained in this work, with its checkpoint selected by validation-set MSE, achieves a per-structure NMAE of 7.7784% with a positive-reference-density mask on the full native grids of all 1,000 MP_EC cubic test structures.

The spectrum enhancement benchmark (Fig. 3e and Supplementary Table S6) uses the HAADF (high-angle annular dark-field) and BF (bright-field) image sets of SFIN to evaluate the image enhancement and detection outputs separately, with the peak signal-to-noise ratio (PSNR) and the structural similarity index (SSIM) as metrics.

The spectrum elucidation benchmark (Fig. 3f and Supplementary Table S7) uses organic molecules with a heavy-atom count (HAC) ≤ 15 from MSD (the Multimodal Spectroscopic Dataset constructed by IBM). The task is to recover molecular graphs from nuclear magnetic resonance (NMR) spectra, and the Top-1/5/10 exact reconstruction accuracy (%) is used as the measure of elucidation accuracy.

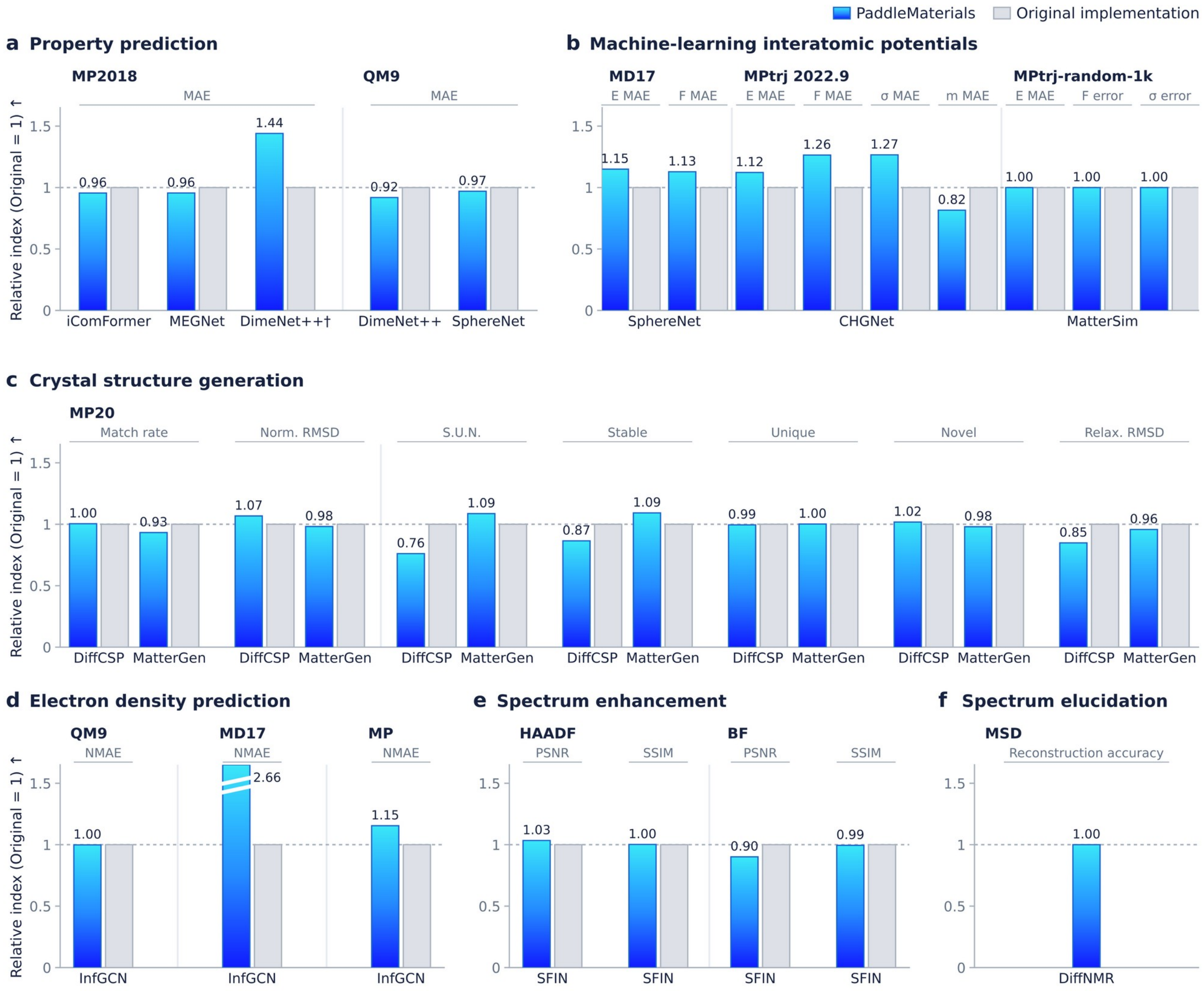

Fig. 3 | Public benchmark results of PaddleMaterials across six task classes. The vertical axes show a relative index: the original-implementation result for the same model, dataset and metric is set to 1, and higher is better. Error metrics use the ratio of the original to the PaddleMaterials result; accuracies, fractions and SSIM use the ratio of the PaddleMaterials to the original result; PSNR is converted from the difference as $10^{\Delta PSNR/10}$; and multiple metrics are combined by an equal-weight geometric mean. a, MEGNet [19], iComFormer [20], DimeNet++ [21] and SphereNet [22], aggregated over the 4 MP2018 or 12 QM9 targets; † the DimeNet++ reference on MP2018 is taken from ref. [23]. b, E, F, σ and m denote energy, force, stress and magnetic moment; SphereNet energy and force errors are aggregated over eight MD17 molecules (force-loss weight p = 100); CHGNet [24] shows the four MAEs separately; MatterSim [11] shows the energy MAE and the L2 errors of force and stress, with the identical results of the 1M and 5M models shown as one group and the original implementation recomputed under the same protocol. c, DiffCSP [25] and MatterGen [2]. Left of the separator is composition-conditioned crystal structure prediction (CSP), which predicts the crystal structure for a given chemical composition and is evaluated by the match rate and normalized RMSD against the true structure; right of the separator is de novo generation, which generates both the composition and the structure: each implementation generated 1,000 structures, which were evaluated with the same MatterSim-assisted workflow for S.U.N. (the fraction of stable, unique and novel structures), Stable, Unique, Novel and the relaxation RMSD (the root-mean-square displacement between a generated structure and its relaxed structure). d, InfGCN [26], with MD17 aggregated over six molecules. e, SFIN [27], with PSNR and SSIM each aggregating the enhancement and detection outputs. f, DiffNMR [28], which exists only as a PaddleMaterials implementation; the metric is the exact-reconstruction accuracy reported in the literature. Full values are given in Supplementary Tables S2–S7.

## CINN compiler backend and cross-model performance evaluation

AI for Materials is moving toward complex tasks that span length and time scales, such as high-accuracy machine-learning interatomic potentials, large-scale molecular dynamics simulations and high-throughput structure generation, and its demand for computing resources is growing rapidly. In Deep Potential molecular dynamics, for example, simulating systems of more than 100 million atoms with ab initio accuracy has already required scaling to the entire Summit supercomputer (about 27,000 GPUs) [29]. At this level of consumption, gains in model execution efficiency translate directly into considerable savings in computing resources and time. Materials-science models typically contain many heterogeneous graph operations, geometric computations and tensor transformations, and they are called repeatedly during simulation, generation, training and inference serving: structure relaxation and molecular dynamics call the potential for energies and forces at every step, and a long simulation typically runs $10^5$–$10^6$ steps; a diffusion model needs about 1,000 denoising steps to generate one structure; high-throughput screening applies the same property model to many candidate structures; training repeats forward and backward computations over a large number of batches; and online inference services keep a resident process that answers prediction requests continuously. The conventional eager mode, that is, immediate dynamic-graph execution, faces serious challenges here. Although eager mode offers excellent flexibility for Python programming and convenient debugging, the extra overhead of scheduling operators one by one, launching kernels frequently and repeatedly reading and writing device memory accumulates sharply over long materials simulations and severely limits the throughput of large GPU clusters.

To break through this computational bottleneck, PaddleMaterials introduces CINN (Compiler Infrastructure for Neural Networks), the PaddlePaddle neural-network compiler, into its underlying architecture as a deeply integrated, optional compiled-execution backend. The core value of CINN lies in exploiting the full potential of the hardware through static compilation: it first captures the numerical computation of the model as a static computational graph, then applies graph optimizations centered on operator fusion, which merges multiple adjacent fine-grained operators into a single kernel and thereby greatly reduces kernel-launch overhead and unnecessary memory traffic of intermediate results, and finally generates code optimized for the target hardware. In PaddleMaterials, CINN compilation happens only at the first call, and subsequent warm calls reuse the compiled result directly. Eager and CINN are therefore two execution paths for the same model: both use the same model definition, weights and configuration, and code that cannot be included in the static graph still runs eagerly; eager execution also serves as the baseline, and the warm speedup is defined as the ratio of the eager warm time to the CINN warm time minus 1, expressed as a percentage. With this design, PaddleMaterials gives materials researchers the best of both worlds: with the same model definition and configuration, and without modifying the upper-level training, prediction or sampling workflows, users can switch to the CINN backend in one step and benefit from the large computational savings of compiled acceleration.

This switch requires no change to the model because PaddleMaterials decouples the scientific computing logic described by a model from the concrete way in which it is executed. To obtain compiled acceleration, conventional approaches often maintain separate eager and static-graph implementations of each model or write dedicated inference wrappers, which leads to duplicated implementations and a long-term maintenance burden. PaddleMaterials manages execution backends, compilation options and caches through a unified orchestrator component and defines stable computation boundaries inside models, such as forward, denoise_step in diffusion models, and graph_encoder and spectrum_encoder for graph/spectrum encoding; the runtime compiles only these numerical boundaries, and the original forward, prediction and sampling protocols of the models remain unchanged. Procedures suited to dynamic graph capture use bytecode-level capture based on SOT (Symbolic OpCode Translator) and allow local graph breaks, whereas procedures that can be made fully static are converted into static programs through abstract syntax tree (AST) transformation. An existing model therefore only needs to inherit the unified runtime capabilities and declare execution boundaries on its compilable numerical functions to use the CINN backend, and neither its network architecture and input/output protocols nor the upper-level Trainer, Predictor and Sampler need to be rewritten. CINN integration thus shifts from "reimplementing a compiled version of the model" to "declaring the compilable boundaries of an existing model", allowing model algorithms and compiler optimization to evolve independently.

The resource savings from this system-level optimization are considerable. The CINN execution path of PaddleMaterials has now been validated against eager execution on 65 pretrained model instances in 11 model groups of 1–16 instances each (Fig. 4b and Supplementary Table S8). In 8 of the groups every model instance runs faster under CINN; the median and maximum warm speedups within these groups are 224% and 336% for SphereNet$^{\text{paddle}}$ MD17 (the maximum, for uracil, is the highest of all model instances), 185% and 216% for MatterSim$^{\text{paddle}}$, 126% and 176% for DimeNet++$^{\text{paddle}}$, 123% and 166% for SFIN$^{\text{paddle}}$, and 47% and 67% for SphereNet$^{\text{paddle}}$ QM9, while CHGNet$^{\text{paddle}}$, DiffCSP$^{\text{paddle}}$ and DiffNMR$^{\text{paddle}}$, with one model instance each, are faster by 107%, 32% and 11%. The gains are most pronounced for the models used in molecular dynamics, the most compute-intensive application: the median speedups of SphereNet$^{\text{paddle}}$ MD17, MatterSim$^{\text{paddle}}$ and CHGNet$^{\text{paddle}}$ are 224%, 185% and 107%, respectively, so the GPU time needed for the same number of calls falls to 31%–48% of that in eager mode. If these speedups carry over to large-scale simulations, calculations that would take months could be completed in weeks, or with fewer than half of the GPUs. Because the benefit comes from reducing the scheduling and memory-access overhead of each call, it accumulates in proportion to the number of calls: molecular dynamics simulations of $10^5$–$10^6$ steps amortize the first-call times of 22–266 s across model groups, and in online inference services the 107% speedup of $\text{CHGNet}^{\text{paddle}}$ means that the same GPU can handle about twice as many requests. The execution path can also be chosen per model, so models with small compilation gains can still be run eagerly. Fig. 4a also compares the eager inference speeds of the original and Paddle implementations on the same inputs as a reference for the eager baseline.

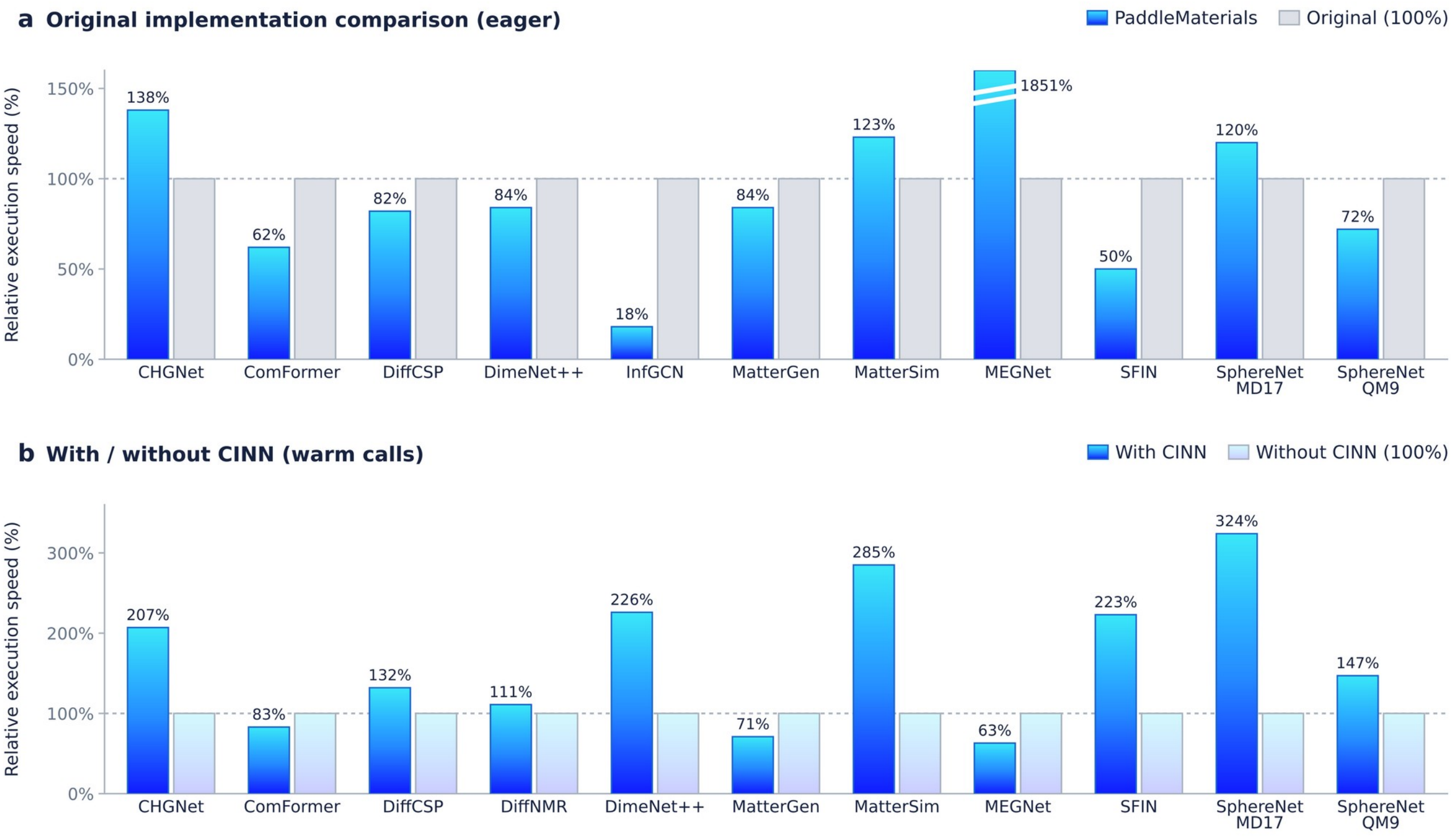

Fig. 4 | Execution performance of PaddleMaterials. The vertical axes show the relative execution speed, that is, the reference time divided by the measured time, expressed as a percentage; each reference is set to 100%, and higher values mean faster execution. A relative speed of R corresponds to "faster by R − 100%" in the text; for example, 207% means faster by 107%. a, Eager inference speed of the original and PaddleMaterials implementations, with the original implementation as the reference; DiffNMR has no original implementation and is not shown. b, Warm-call speed of PaddleMaterials models without and with CINN, with CINN disabled (eager execution) as the reference; first-call compilation is excluded, groups with several model instances show the median over their instances, and InfGCN is not included in the CINN tests. Panels a and b use different test inputs; the times, speedup ranges and first-call times of each model group are given in Supplementary Table S8.

## Cross-task computation and application validation

Previous studies have demonstrated applications of PaddleMaterials and its associated models across different material systems and research tasks. In two-dimensional (2D) materials research, researchers combined $\text{CHGNet}^{\text{paddle}}$ in the Paddle framework with DFT to carry out structure relaxation and stability screening through active learning, built ML2DDB, a monolayer 2D materials database containing 242,546 DFT-validated structures, and used $\text{MatterGen}^{\text{paddle}}$, implemented in the Paddle framework, for conditional structure generation [30]. In research on layered double hydroxide (LDH)-supported single-atom catalysts, researchers used the $\text{iComFormer}^{\text{paddle}}$ model provided by PaddleMaterials and combined geometric perturbation, two-stage model training and DFT calculations to screen microenvironments for the hydrogen evolution reaction, identifying the top 500 ranked candidates among 4,700 unexplored structures [31]. In research on metal-complex catalysts, researchers used $\text{InfERN}^{\text{paddle}}$, a model in PaddleMaterials based on a Graphon neural operator, to predict the three-dimensional (3D) electron densities of metal coordination compounds through rotation- and translation-equivariant modeling; the study reports that the predicted electron-density distributions agree with quantum-chemical calculations and that the prediction time is reduced relative to DFT calculations [32]. These cases cover material systems including 2D crystals, LDH-supported single-atom catalysts and metal coordination compounds, and show how the platform and its associated models support research tasks such as structure exploration, catalytic microenvironment screening and electron-density prediction. This section uses the crystalline material $LiMnO_2$ and the organic molecule toluene to test whether the platform can link models of different tasks into cross-task workflows, and to assess whether the accuracy of the workflow results can support application studies and whether their computational cost can support large-scale screening.

Both cases follow the same workflow (Fig. 5): structure generation or spectral elucidation gives the initial configuration $G_0$, structure relaxation with a machine-learning interatomic potential gives the relaxed configuration $G_{relax}$, scalar properties and electron densities are then predicted on both configurations, and same-geometry DFT single-point calculations and a DFT geometry optimization starting from $G_{relax}$ serve as external references; the computational settings of each step are given in Methods. $LiMnO_2$ is a nickel- and cobalt-free candidate cathode material for lithium-ion batteries, and local coordination and the Jahn–Teller distortion of $Mn^{3+}$ are important factors in its structural study [33, 34]. This case starts from a composition condition: DiffCSP$^{paddle}$ generates an eight-atom $Li_2Mn_2O_4$ candidate crystal for Li:Mn:O = 1:1:2, which serves as $G_0$ after structure checks and is then relaxed with the CHGNet$^{paddle}$ potential, first with a fixed cell and then with a variable cell (Fig. 5a). The toluene ($C_7H_8$) case starts from spectroscopic information: DiffNMR$^{paddle}$ generates candidate molecular graphs conditioned on a literature NMR peak list and the heavy-atom count, and one recovered graph that exactly matches toluene is selected; hydrogen addition with RDKit and ETKDGv3 embedding give $G_0$, which is then relaxed with the SphereNet$^{paddle}$ MD17 potential (Fig. 5b).

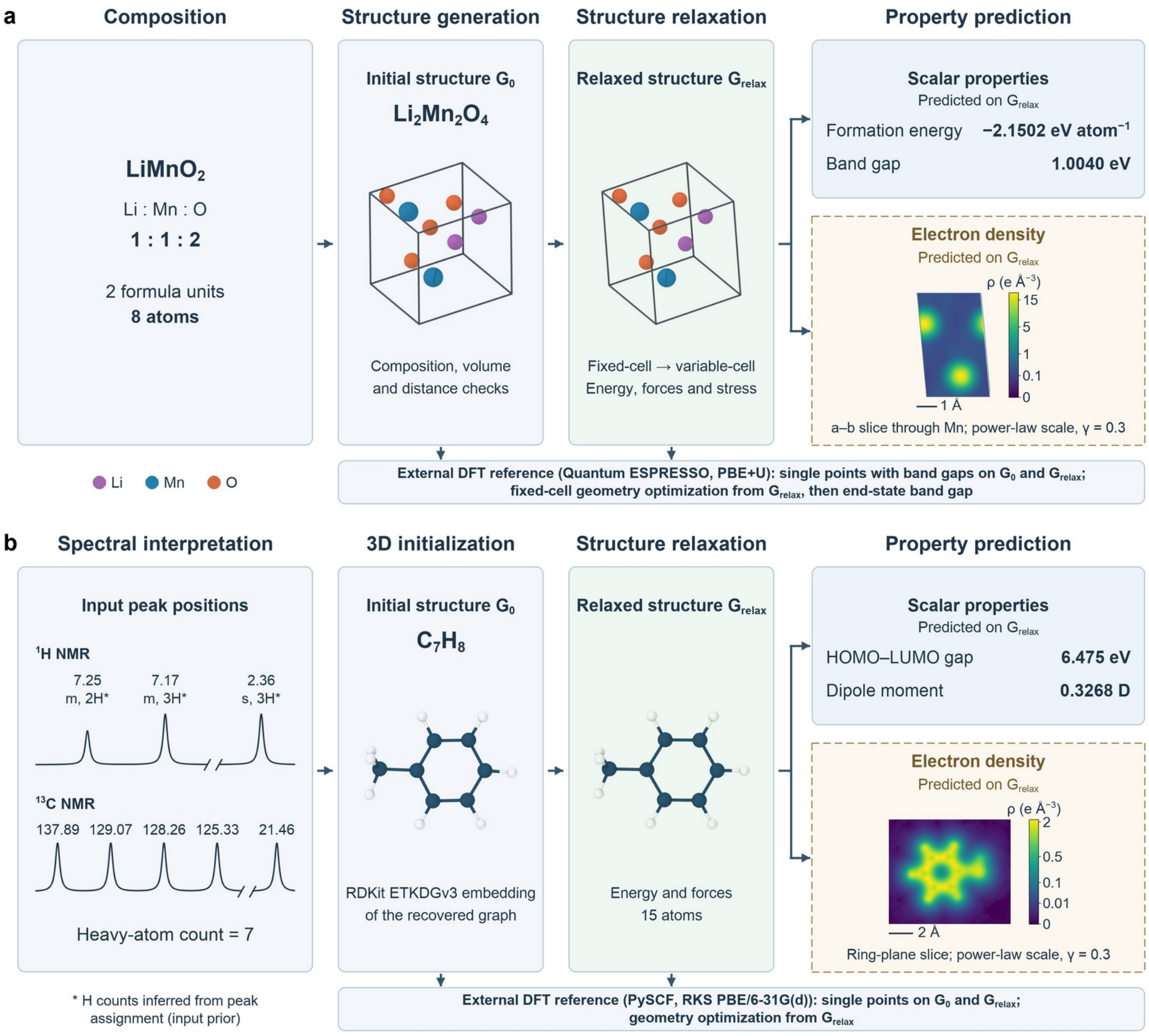

Fig. 5 | Cross-task computational workflows for $LiMnO_2$ and toluene. In both workflows, structure generation or spectral elucidation gives the initial configuration $G_0$, structure relaxation gives the relaxed configuration $G_{relax}$, and properties and electron densities are then predicted on $G_{relax}$; the bottom bars show the external DFT reference calculations. The density insets are native-grid slices of the InfGCN$^{paddle}$ prediction on $G_{relax}$. The color scale uses the viridis colormap with a power-law scale ($\gamma = 0.3$), in units of e Å$^{-3}$; for display, negative values are clipped to 0, the upper limit is set to the maximum of the full predicted field, and values are linearly interpolated between grid points. **a**, $LiMnO_2$. With Li:Mn:O = 1:1:2 as the composition condition (two formula units, eight atoms in total), DiffCSP$^{paddle}$ generates a $Li_2Mn_2O_4$ candidate crystal, which, after checks of

composition, volume per atom and minimum interatomic distance, serves as $G_0$; CHGNet$^{paddle}$ relaxes $G_0$ first with a fixed cell and then with a variable cell. The property box gives the formation energy and band gap predicted by iComFormer$^{paddle}$. The density slice is the a–b plane containing two Mn atoms (fractional coordinate c = 67/108). Scale bar, 1 Å. Bottom bar: Quantum ESPRESSO performs same-geometry PBE+U single-point calculations on $G_0$ and $G_{relax}$, giving k-point-sampled band gaps, and then a fixed-cell PBE+U geometry optimization starting from $G_{relax}$, followed by a band-gap calculation on the optimized structure (Table 2). **b**, Toluene. The inputs are the literature $^1$H and $^{13}$C NMR peak positions in $CDCl_3$ ([35]) and the heavy-atom count; no molecular formula is given. The spectra in the left column are schematics drawn from the input peak list (peak spacing not to scale; $^1$H peak heights follow the hydrogen counts). An asterisk (*) after the hydrogen count of a $^1$H peak marks a hydrogen count inferred from the peak assignment. DiffNMR$^{paddle}$ recovers the molecular graph, hydrogen addition with RDKit and ETKDGv3 embedding give $G_0$, and $G_0$ is relaxed with the SphereNet$^{paddle}$ MD17 potential. The property box gives the HOMO–LUMO gap (that is, the molecular gap) and dipole moment predicted by the SphereNet$^{paddle}$ QM9 property model; results for $G_0$, $G_{relax}$ and the DFT-optimized structure are given in Table 4. The density slice is the xy plane through the benzene ring (all seven carbon atoms lie within ±0.16 Å of this plane). Scale bar, 2 Å. Bottom bar: PySCF performs same-geometry RKS PBE/6-31G(d) single-point calculations on $G_0$ and $G_{relax}$ and then a PBE geometry optimization starting from $G_{relax}$ (Table 4).

In terms of accuracy, the formation energies predicted by the models in the $LiMnO_2$ case are close to the same-geometry DFT values, and the configuration obtained by structure relaxation is close to the DFT-optimized structure, whereas the band gaps deviate more (Table 2). The same-protocol PBE+U formation energies are −2.1167, −2.1566 and −2.1576 eV atom$^{-1}$ for $G_0$, $G_{relax}$ and the DFT-optimized structure, respectively; MEGNetPlus$^{paddle}$ and iComFormer$^{paddle}$ predict −2.1231 and −2.1163 eV atom$^{-1}$ on $G_0$ and −2.1588 and −2.1502 eV atom$^{-1}$ on $G_{relax}$, differing from the same-geometry DFT values by no more than 7 meV atom$^{-1}$; on the DFT-optimized structure they predict −2.1468 and −2.1532 eV atom$^{-1}$, differing by 10.8 and 4.4 meV atom$^{-1}$, respectively. The PBE+U energies of $G_0$ and $G_{relax}$ are 40.9 and 1.0 meV atom$^{-1}$ higher than that of the optimized structure, and the root-mean-square (RMS) atomic displacement after DFT optimization is 0.0253 Å, so the energy decrease is achieved mainly in the structure relaxation step and $G_{relax}$ is already close to the DFT-optimized structure. For the band gap, both models reproduce the opening of the gap upon structure relaxation but overestimate its value. On $G_0$, the two models predict 0.0241 and 0.1425 eV, whereas in the same-geometry PBE+U single-point calculation the Fermi level crosses the bands in the spin-up channel and $G_0$ is metallic (k-point-sampled gap of about 0.003 eV), which is qualitatively consistent with the near-zero gaps given by the models; on $G_{relax}$, the two models predict 1.1277 and 1.0040 eV, 0.28–0.41 eV higher than the PBE+U value of 0.7196 eV. The PBE+U band gap of the optimized structure is 0.7942 eV, 0.0746 eV higher than that of $G_{relax}$; compared with the formation-energy change of only 1.0 meV atom$^{-1}$, the band gap is more sensitive to this small structural adjustment. On the optimized structure, the two models predict 1.2678 and 0.9774 eV, still 0.18–0.47 eV higher than PBE+U. The formation energies and relaxed structures can therefore be used directly for initial candidate screening, whereas structure-sensitive electronic properties such as the band gap still require DFT verification.

Table 2 | Property predictions and DFT comparison for the eight-atom $LiMnO_2$ case.

| Quantity | Initial configuration $G_0$ | Relaxed configuration $G_{relax}$ | DFT-optimized structure |
|---|---|---|---|
| **Formation energy (eV atom$^{-1}$)** | | | |
| MEGNetPlus$^{paddle}$ | −2.1231 | −2.1588 | −2.1468 |
| iComFormer$^{paddle}$ | −2.1163 | −2.1502 | −2.1532 |
| PBE+U (same geometry) | −2.1167 | −2.1566 | −2.1576 |
| **Band gap (eV)** | | | |
| MEGNetPlus$^{paddle}$ | 0.0241 | 1.1277 | 1.2678 |
| iComFormer$^{paddle}$ | 0.1425 | 1.0040 | 0.9774 |
| PBE+U (same geometry) | 0.003 (metallic) | 0.7196 | 0.7942 |
| **Structure (PBE+U)** | | | |
| Energy relative to the optimized structure (meV atom$^{-1}$) | 40.9 | 1.0 | 0 |
| Maximum atomic force (eV Å$^{-1}$) | 0.4848[a] | 0.1318 | 0.00603 |
| Atomic displacement, RMS / maximum (Å) | — | 0 (reference) | 0.0253 / 0.0364 |

Note: Model predictions and PBE+U references in the same column use the same geometry. PBE+U formation energies are referenced to bcc Li, α-Mn and $O_2$; band gaps are Kohn–Sham gaps sampled on a 14×12×12 k-point grid. The energy relative to the optimized structure is the difference in

same-protocol PBE+U total energy per atom, which equals the difference in formation energy. Maximum atomic forces are computed with DFT, and atomic displacements are measured from $G_{relax}$; the results are limited to the fixed cell and the selected magnetic branch. [a] Single-point calculation on an 8×6×6 k-point grid; the other force values use a 12×10×10 grid. "—" denotes not computed.

In the toluene case, the accuracy of the workflow depends first on whether spectral elucidation recovers the correct molecular graph. With the integrals unknown, none of the 16 candidates exactly matched toluene; after the 2H/3H/3H integrals at δ 7.25/7.17/2.36 ppm inferred from the peak assignment were added, 2 of the 16 candidates matched exactly (Table 3), showing that this step depends on prior information such as the integrals. Because the integrals come from the known peak assignment and the candidate used downstream was selected according to the known molecular graph of toluene, this case is a workflow validation conditioned on a matched candidate and does not constitute a blind test on an unknown molecule. Given the correct molecular graph, structure relaxation accomplishes most of the geometric adjustment: the PBE energy decreases by 0.2219 eV in total from $G_0$ to the DFT-optimized structure, about 97% of which is already achieved in the structure relaxation, and a PBE geometry optimization starting from $G_{relax}$ converged in 12 steps to a maximum atomic force of 0.00710 eV Å$^{-1}$, with an RMS atomic displacement of 0.0088 Å from $G_{relax}$. For the properties, the SphereNet$^{paddle}$ QM9 property model and the PBE reference follow different theoretical protocols, so their absolute values are not compared as same-protocol errors, but they change in the same direction upon structure relaxation: the dipole moments from the model and from PBE increase by about 0.10 and 0.11 D, respectively, and both molecular gaps decrease (Table 4). From $G_{relax}$ to the optimized structure, the PBE gap and dipole moment change by 0.031 eV and 0.002 D, and the corresponding model predictions change by 0.025 eV and less than 0.001 D, so for this molecule $G_{relax}$ can be used directly for property prediction. In both cases, InfGCN$^{paddle}$ outputs electron densities on $G_0$ and $G_{relax}$, and Fig. 5a, b shows the predicted slices for $G_{relax}$; for toluene, the signed integral of the raw predicted field of $G_{relax}$ over the full sampling grid is 35.98 e, close to the 36 e obtained from the valence-electron count.

Table 3 | Candidate statistics for the spectroscopic recovery of toluene.

| Run / integral condition | Number of candidates | RDKit-valid and connected | Exact graph match to the target molecule |
|---|---|---|---|
| Toluene / integrals unknown | 16 | 5 | 0/16 |
| Toluene / 2H/3H/3H integrals at δ 7.25/7.17/2.36 ppm inferred from peak assignment | 16 | 10 | 2/16 |

Note: Exact graph matching excludes stereochemistry, and invalid and disconnected candidates are counted in the denominator. The integrals were inferred from the known peak assignment; 2/16 is the match ratio within this batch of candidates.

Table 4 | Molecular property predictions and DFT comparison for the toluene case.

| Quantity | Initial configuration $G_0$ | Relaxed configuration $G_{relax}$ | DFT-optimized structure |
|---|---|---|---|
| **Molecular gap (eV)** | | | |
| SphereNet$^{paddle}$ | 6.613 | 6.475 | 6.450 |
| PBE (same geometry) | 5.003 | 4.958 | 4.927 |
| **Dipole moment (D)** | | | |
| SphereNet$^{paddle}$ | 0.2242 | 0.3268 | 0.3271 |
| PBE (same geometry) | 0.2580 | 0.3655 | 0.3635 |
| **Structure (PBE)** | | | |
| Energy relative to the optimized structure (eV) | 0.2219 | 0.0056 | 0 |
| Maximum atomic force (eV Å$^{-1}$) | 1.640 | 0.1440 | 0.00710 |
| Atomic displacement, RMS / maximum (Å) | — | 0 (reference) | 0.0088 / 0.0147 |

Note: Model predictions and PBE references in the same column use the same geometry; PBE denotes RKS PBE/6-31G(d) calculations. The model molecular gap is predicted directly by the QM9 gap task, whereas the PBE gap is LUMO−HOMO from the same single-point calculation; because the two follow different theoretical protocols, their difference is not treated as a same-protocol error. The DFT-optimized structure was obtained with ASE BFGS starting from $G_{relax}$ until the maximum atomic force fell below 0.01 eV Å$^{-1}$; maximum atomic forces are 3D force-vector norms, and atomic displacements are measured from $G_{relax}$. "—" denotes not computed.

In terms of cost, the model steps in the $LiMnO_2$ workflow take far less time than the DFT calculations. CHGNet$^{paddle}$ completed the fixed-cell and variable-cell relaxations of two candidate structures in 111 s in total, single property predictions with MEGNetPlus$^{paddle}$ and iComFormer$^{paddle}$ took 0.12–0.23 s, and InfGCN$^{paddle}$ predicted the electron density on 746,496 grid points in 3.6 s, whereas one same-geometry PBE+U single-point calculation for the same eight-atom structure required about 7.4–11.9 CPU core-hours, and the fixed-cell DFT optimization starting from $G_{relax}$ involved another 8 self-consistent field (SCF) calculations. In core-hours, the total cost of the model steps for relaxation, property prediction and density prediction is less than 1/20 of that of one DFT single-point calculation. For a small molecule such as toluene, a PBE/6-31G(d) single-point calculation takes only about 30 s, and the time advantage of the models is correspondingly smaller.

The two cases show that PaddleMaterials can link structure generation or spectral elucidation, structure relaxation, property prediction and electron-density prediction for a crystalline material and an organic molecule into cross-task workflows through explicit crystal structures, molecular graphs and geometric states, and can be used together with external tools such as RDKit, Quantum ESPRESSO and PySCF. The accuracy of the workflow results is limited but sufficient to support application studies: the formation energies and relaxed structures are close to the DFT references, and the molecular properties change in the same direction as in DFT upon structure relaxation, so these results can be used for initial candidate screening and structural preprocessing; the band-gap values still deviate, and spectral elucidation depends on prior information such as the integrals, so these steps require verification by DFT calculations or experimental information. Because the model steps cost far less than crystal DFT calculations, they are suited to large-scale screening, leaving DFT for the accurate verification of a few candidates; the studies on 2D materials and single-atom catalysts described above likewise combined model-based screening with DFT calculations [30, 31].

# Discussion

This work addresses the difficulty of organizing heterogeneous machine-learning tasks in materials science within a single software system by building and evaluating PaddleMaterials. Starting from a unified formal description of tasks, we derived the design requirements of the platform from the differences among six task classes in input representations, computational structures and training objectives, modes of use, output semantics and execution efficiency, presented a three-layer architecture that meets these requirements, and evaluated the design with public benchmarks, CINN compiled-execution experiments and two cross-task case studies. The results support the following design principle: a unified platform for materials AI need not compress different tasks into a single representation; instead, the platform can share data processing, training, evaluation, model management and execution, while each task explicitly defines its representations, model architectures, training objectives and scientific semantics, which remain verifiable at task boundaries. This design lowers the cost of organizing materials AI research at both the model and the workflow levels. At the model level, single-file model organization, the registry mechanism and configuration-driven construction confine the work of adding or migrating a model mainly to the model's own computational definition: once a model provides the agreed forward(), predict() or sample() interfaces, it can reuse the training, prediction, sampling and simulation workflows organized by Trainer, Predictor, Sampler and Calculator, and integrating CINN only requires declaring compilable computation boundaries instead of maintaining a separate compiled version. At the workflow level, tasks record physical quantities, units, aggregation levels and reference conventions at their input and output boundaries, and pretrained-model packages store weights together with configurations, normalization statistics, vocabularies and other states. Downstream tasks can therefore determine whether an upstream output can be used directly or which conversion it requires, and cross-task comparisons presuppose explicit computational protocols.

Taken together, the benchmark, case-study and compiled-execution results show the value of PaddleMaterials for AI for Materials research in four respects. First, model coverage and accuracy alignment: the 11 migrated models for six task classes were re-evaluated on public benchmarks within one code base and generally retained the accuracy reported for the original implementations, so that models from different tasks and sources can be reproduced, compared and invoked through consistent data processing, training and evaluation workflows; these results also provide a reference for cross-framework reproduction. Second, model reuse: the platform distributes weights together with their data-processing states as pretrained-model packages, so researchers can apply existing models directly to new structures, simulation conditions and characterization data without retraining or rebuilding preprocessing pipelines. Third, workflow integration and computational cost: through interoperability with tools such as pymatgen, RDKit and ASE, structure generation, spectrum elucidation, relaxation and property prediction can be chained into continuous workflows that connect to external calculations such as DFT. The $LiMnO_2$ case generates crystal candidates from a composition condition and the toluene case recovers molecular graphs from spectroscopic information; both use atomic structures as the common input for subsequent relaxation and property calculations, and the scalar-property and electron-density outputs provide global and spatially resolved descriptions, respectively. In the $LiMnO_2$ case, the two property models predict a decrease in formation energy of 34–36 meV $atom^{-1}$ from $G_0$ to $G_{relax}$, close to the 40 meV $atom^{-1}$ obtained with PBE+U; subsequent DFT optimization lowers the energy of $G_{relax}$ by only about 1 meV $atom^{-1}$ (Table 2), and the total cost of the model steps is less than 1/20 of one DFT single-point calculation. This reflects a division of labor between models and first-principles calculations: models handle large-scale prescreening and structure preprocessing, and DFT is reserved for the accurate verification of a few candidates. Fourth, execution efficiency: the CINN compiled-execution backend optimizes execution efficiency uniformly without changing model definitions. The eager path retains the flexibility needed for model development and debugging and serves as the performance baseline, whereas CINN, through computational-graph optimization, operator fusion and code generation for the target hardware, provides a faster execution path for computations such as molecular dynamics and diffusion sampling that call the same model repeatedly. This benefit comes at the cost of first-call compilation: in Supplementary Table S8, the group medians of the first-call time range from 22 to 266 s. For $CHGNet^{paddle}$, each warm call is about 20 ms faster than under eager execution, so about 7,000 calls are enough to offset the first-call compilation time; for long simulations with far more steps, the compilation cost is fully amortized.

Future work will proceed in three directions. The first is to extend tasks and systems to cover more material classes and data modalities. The second is to strengthen cross-task evaluation and execution optimization and to extend the effective coverage of CINN. The third is to support automated research workflows by wrapping models as standardized tools for high-throughput screening, active learning and large-language-model agents, allowing PaddleMaterials to develop from a collection of models into reusable infrastructure for materials research.

# Methods

## Code, model and data versions, and performance evaluation

The experiments use PaddleMaterials (https://github.com/PaddlePaddle/PaddleMaterials) and the public configuration for each task, with PaddlePaddle 3.3.1, Python 3.12.14 and NumPy 1.26.4 as the main runtime environment and the corresponding versions of Paddle, ASE, PySCF or Quantum ESPRESSO for the application calculations of each task; input structures, model outputs, calculation scripts and result tables are provided with the reproducibility materials. Results from different frameworks are compared only when the data splits, structural states, target definitions and units are consistent. The CINN tests record the first-call time, which includes compilation, and the warm time, the median of the 10 calls after the first call; in the eager comparison, the original and Paddle implementations are timed alternately in separate processes for 4 rounds, with 30 timed calls after warm-up in each round, and times are the median of the round medians, with speedups derived from the geometric mean of the per-round time ratios. Both types of test were run on NVIDIA A100-SXM4-40GB GPUs; timings exclude model and weight loading, and the GPU is synchronized before and after each call.

## $LiMnO_2$ structure generation, relaxation and property prediction

$DiffCSP^{paddle}$ generates candidate structures conditioned on the element counts of $Li_2Mn_2O_4$, using 1,000 diffusion steps. After the candidates are checked for a finite cell, volume per atom and minimum interatomic distance, one eight-atom structure is selected as the initial configuration $G_0$; fixed-cell and variable-cell relaxation with $CHGNet^{paddle}$ gives the relaxed configuration $G_{relax}$, and the cell, coordinates, energy, forces and stress are saved. The external reference calculations are performed with Quantum ESPRESSO (QE) using the spin-polarized plane-wave PBE+U method (the PBE exchange–correlation functional with Hubbard U = 3.9 eV applied to the Mn 3d orbitals) and GBRV pseudopotentials, in two steps: same-geometry single-point calculations on $G_0$ and $G_{relax}$ give formation energies and k-point-sampled Kohn–Sham band gaps, and a fixed-cell optimization starting from $G_{relax}$ then assesses further structural adjustment through atomic forces and displacements, after which the formation energy and band gap of the optimized structure are calculated. Calculation inputs, outputs and pseudopotential files are provided with the reproducibility materials. The same-geometry band-gap single-point calculation and the subsequent fixed-cell optimization each use their own k-point sampling and smearing settings, and total energies from the two protocols are not mixed. Formation energies are computed as $E_f = (E_{cell} - N\mu_{Li} - N\mu_{Mn} - 2N\mu_O)/(4N)$ (N = 2 for the eight-atom cell); all phases use the above PBE+U settings, 95/1140 Ry cutoffs and 0.001 Ry Marzari–Vanderbilt smearing, and the total energies include the smearing term. The chemical potentials $\mu_{Li}$, $\mu_{Mn}$ and $\mu_O$ are taken from calculations of bcc Li, 58-atom α-Mn (also with Mn U = 3.9 eV) and triplet $O_2$, respectively. Structural differences are computed in the same cell with the same atom order, taking periodic boundaries into account; the maximum atomic force is taken from the norms of the 3D force vectors. MEGNetPlus$^{paddle}$ and iComFormer$^{paddle}$ each output formation energy and band gap, using units and inverse transforms consistent with the training targets. InfGCN$^{paddle}$ predicts the electron density on $G_0$ and on $G_{relax}$, with both predictions using the same checkpoint and the same fractional-coordinate grid; the prediction on $G_0$ is shown alongside the same-geometry QE/PBE+U reference (Supplementary Fig. S1), and the prediction on $G_{relax}$ is shown for qualitative illustration only (Fig. 5a). For the timings, one same-geometry PBE+U single-point calculation requires about 7.4 and 11.9 CPU core-hours with 12×10×10 and 14×12×12 k-point meshes, respectively; the CHGNet$^{paddle}$ relaxations use 2 CPU threads, and their timing includes model loading and derivative checks; the MEGNetPlus$^{paddle}$ and iComFormer$^{paddle}$ property predictions use a single CPU core; and the InfGCN$^{paddle}$ density prediction uses one A100 GPU. The relationships between the states, the DFT settings and the scope of the density-target checks are detailed in the Appendix.

### Toluene structure, spectral elucidation and molecular properties

DiffNMR$^{paddle}$ generates molecular graphs conditioned on the NMR peak list and the heavy-atom count, with the integral conditions inferred from the literature peak assignment. After the candidates are checked for RDKit validity, connectivity and exact graph match to the known target, one of the exactly matched candidates is selected. Hydrogen addition and ETKDGv3 embedding of the recovered graph give the initial configuration $G_0$, which then undergoes structure relaxation with the SphereNet$^{paddle}$ MD17 potential checkpoint to give the relaxed configuration $G_{relax}$. The selected configuration passes a local force–energy consistency check. Molecular properties are predicted separately with the QM9 property checkpoints. The PySCF reference uses neutral closed-shell RKS PBE/6-31G(d), and non-periodic geometry optimization uses ASE BFGS. InfGCN$^{paddle}$ predicts the electron density point by point on the sampling grids of $G_0$ and $G_{relax}$, and integrals use the unclipped, unrescaled raw signed output; the same-geometry all-electron reference serves separately as a cross-representation comparison. Inputs, model usage, optimization conditions and grid settings are given in the Appendix.

# Appendix

## $LiMnO_2$ case

In Fig. 5a and Table 2 of the main text, the eight-atom candidate generated by $\text{DiffCSP}^{\text{paddle}}$ is the initial configuration $G_0$; fixed-cell and variable-cell relaxation with $\text{CHGNet}^{\text{paddle}}$ gives the relaxed configuration $G_{\text{relax}}$; and a fixed-cell QE/PBE+U optimization starting from $G_{\text{relax}}$ then gives the DFT-optimized structure. Same-geometry single-point calculations are performed on $G_0$ and on $G_{\text{relax}}$, and all three states keep the same elements and atom order.

The band-gap calculations use collinear spin-polarized PBE+U (U = 3.9 eV on the Mn 3d orbitals), GBRV pseudopotentials and 95/1140 Ry cutoffs, with a 14×12×12 k-point grid and 0.001 Ry Marzari–Vanderbilt smearing. Each spin channel includes 36 bands; for $G_{\text{relax}}$, the spin-up and spin-down channels contain 34 and 26 occupied bands, respectively, and the band gap is determined from the eigenvalues of the occupied and unoccupied states in both channels. Reducing the smearing from 0.002 to 0.001 Ry changes the band gap by $4.06\times10^{-6}$ eV, and refining the k-point grid from 12×10×10 to 14×12×12 changes it by 0.00790 eV. The same-geometry calculation for $G_0$ uses the same settings with 48 bands, so that the highest band is unoccupied at all k-points; the number of occupied bands in its spin-up channel varies between 33 and 36 across k-points, so $G_0$ is metallic.

The fixed-cell optimization starts from $G_{\text{relax}}$ and uses spin-polarized QE/PBE+U (Mn U = 3.9 eV) with GBRV pseudopotentials (version 1.4 for Li, 1.5 for Mn and 1.2 for O), 95/1140 Ry cutoffs, a 12×10×10 k-point grid and 0.01 Ry smearing; the SCF convergence threshold is $10^{-9}$ Ry, the ionic energy threshold is $10^{-5}$ Ry and the force-component threshold is $5\times10^{-4}$ Ry $\text{bohr}^{-1}$, and the cell and the selected magnetic branch are kept unchanged. The optimization involved 8 SCF calculations and 7 BFGS steps in total, and the maximum atomic force decreased from 0.13183 to 0.00603 eV $\text{Å}^{-1}$. With atoms matched by element and atom index and distances taken as the shortest periodic distances in the same cell, the RMS displacement of the eight atoms from $G_{\text{relax}}$ is 0.02527 Å and the maximum displacement is 0.03645 Å; the four O atoms are displaced by 0.03478–0.03645 Å, and all Li and Mn atoms by less than 0.00141 Å.

Electron densities are predicted with the $\text{InfGCN}^{\text{paddle}}$ MP_EC checkpoint evaluated in Supplementary Table S5, and the MP_EC density labels come from VASP/PAW Opt-B88vdW calculations. The predictions on $G_0$ and $G_{\text{relax}}$ use the same checkpoint and the same 64×108×108 fractional-coordinate grid (746,496 points), and the statistics are computed from the unclipped, unrescaled raw signed output. Supplementary Fig. S1 shows the prediction on $G_0$ alongside the same-geometry QE/PBE+U/GBRV pseudo-valence electron density; the two differ in computational protocol and density representation, so the comparison serves only as a cross-method reference: the signed integrals of the prediction and the reference are 56.16 and 60.00 e, respectively, the relative L1 field difference is 17.99%, and 5.65% of the grid points in the prediction are negative. For the prediction on $G_{\text{relax}}$ (Fig. 5a of the main text), the signed integral is 56.60 e, 4.57% of the grid points are negative, and the density ranges from −0.196 to 18.93 e $\text{Å}^{-3}$.

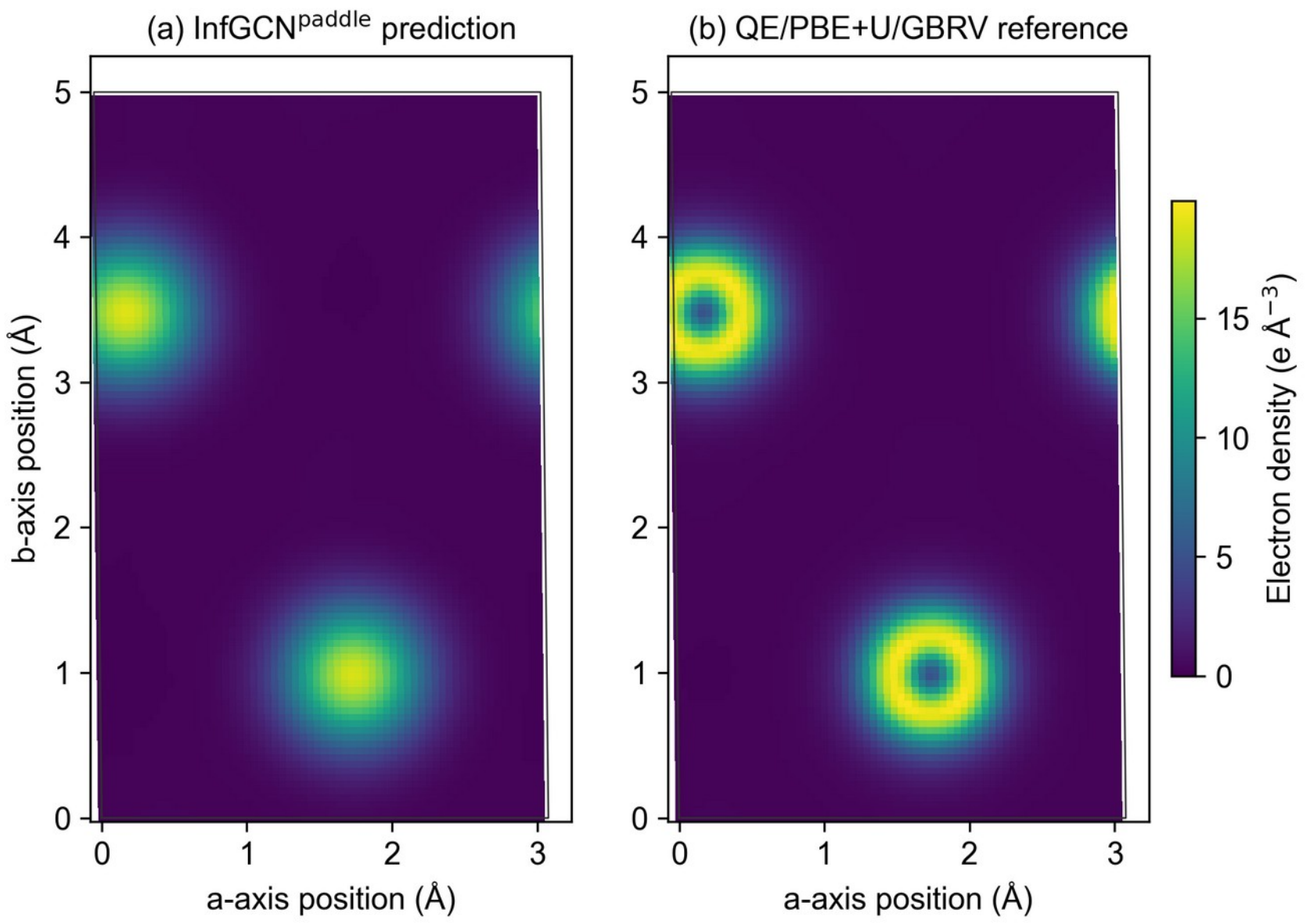

Supplementary Fig. S1 | Electron density of the eight-atom $LiMnO_2$ initial configuration $G_0$. a, InfGCN$^{paddle}$ prediction; b, same-geometry QE/PBE+U/GBRV reference. Both panels show the a–b slice at index 67 along the c direction of the 64×108×108 grid (fractional coordinate c = 67/108, the same plane as in Fig. 5a of the main text), with the same linear color scale in units of e Å$^{-3}$.

## Toluene case

The inputs are the $^{1}$H NMR peaks (δ 7.25, m; 7.17, m; 2.36, s) and $^{13}$C NMR peaks (δ 137.89, 129.07, 128.26, 125.33, 21.46) of toluene in $CDCl_3$ reported in ref. [35], together with a heavy-atom count of 7; no molecular formula is given, and the 2H/3H/3H integrals of the $^{1}$H peaks are inferred from the peak assignment. DiffNMR$^{paddle}$ generates 16 candidates under each of two conditions, with the integrals unknown and with the inferred integrals given (Table 3 of the main text); the subsequent calculations use one recovered graph that exactly matches toluene, and hydrogen addition with RDKit and ETKDGv3 embedding give $G_0$. Structure relaxation uses the SphereNet$^{paddle}$ MD17 potential checkpoint, the molecular gap and dipole moment are predicted with the SphereNet$^{paddle}$ QM9 property checkpoints, and the electron density is predicted with the InfGCN$^{paddle}$ QM9 checkpoint, whose training targets are PAW/PBE valence electron densities; the SphereNet$^{paddle}$ MD17 potential checkpoint and QM9 property checkpoints all use the adjusted torsion-angle treatment.

Structure relaxation uses BFGS, and the potential is evaluated in float64 precision. The configuration relaxed with the original float32 implementation failed the force–energy consistency check: for the 6 force components checked at h = $10^{-3}$ and $10^{-4}$ Å, the maximum force differences were 0.669 and 3.11 eV Å$^{-1}$, respectively. After the torsion-angle treatment was adjusted and float64 was adopted, the configuration used here reached the model force threshold in 15 BFGS steps, with a maximum atomic force of 0.018975 eV Å$^{-1}$; finite-difference checks of all 45 force components gave maximum errors of $8.78\times10^{-7}$ and $7.00\times10^{-8}$ eV Å$^{-1}$ at h = $10^{-5}$ and $10^{-6}$ Å, respectively.

The PBE reference uses PySCF RKS PBE/6-31G(d) for the neutral singlet state, with integration grid level 4 and an SCF convergence threshold of $10^{-10}$ Ha. The DFT-optimized structure was obtained with ASE BFGS starting from $G_{relax}$ (maxstep 0.1 Å, maximum atomic force threshold 0.01 eV Å$^{-1}$) after a total of 12 BFGS steps and 13 SCF and gradient calculations, and the maximum atomic force of the optimized structure is 0.0071 eV Å$^{-1}$.

The electron density of toluene is predicted point by point for $G_{relax}$ on a 90×83×69 sampling grid (spacing 0.2 Å, 515,430 points in total). The integral of the raw signed output over the full grid is 35.98 e, and the predicted density ranges from −0.0006 to 2.158 e Å$^{-3}$. The slice in Fig. 5b of the main text is the xy grid plane closest to the median z coordinate of the carbon atoms (z = 0.106 Å). The same-geometry B3LYPG/6-31G(2df,p) all-electron density serves only as a cross-representation comparison and is not used for error evaluation.

## General abstractions of heterogeneous tasks

Supplementary Table S1 | General abstractions of heterogeneous materials-science tasks and example implementations in PaddleMaterials.

| Task family | Computational abstraction | Learning and modeling paradigm | Representations and inputs/outputs | Concrete materials-science instances | Task specificity | Main inductive biases and physical constraints | Representative models | Representative datasets | Evaluation metrics |
|---|---|---|---|---|---|---|---|---|---|
| Property prediction | Materials representation → target property;<br>$y = f_\theta(X)$ | Supervised regression or classification | Input: composition, structure or their descriptors<br>Output: scalar, vector or tensor properties, or classes. | Property estimation, materials screening and structure–property relationship modeling. | Target definition, reference method, units, output type and target transformation. | Permutation and geometric symmetries consistent with the output type; periodicity; scale dependence consistent with the target. | MEGNet; iComFormer; DimeNet++; SphereNet | MP2018/MP2024; JARVIS; QM9; OMat24; OMol25 | Regression: MAE, RMSE, $R^2$;<br>Classification: accuracy, F1, AUROC |
| Structure generation | Structure distribution modeling;<br>$X \sim p_\theta(X)$ or $X \sim p_\theta(X \mid c)$ | Generative learning; conditional or unconditional generation | Input: conditions such as composition or target properties (if any).<br>Output: molecular graphs or 3D atomic structures; periodic structures also include the lattice. | Candidate structure generation, structure prediction and inverse design. | Composition constraints, structure representation, periodicity, condition definition and sampling strategy. | Permutation and geometric symmetries; chemical validity; boundary conditions for periodic systems. | DiffCSP; MatterGen | MP20; Alex-MP20; ML2DDB | Generation quality: validity, uniqueness, novelty, condition satisfaction rate.<br>Structure prediction: match rate, RMSD |
| Machine-learning interatomic potential (MLIP) | Atomic structure → energy and forces;<br>$X \mapsto (E, \mathbf{F})$<br>Energy-based models:<br>$\mathbf{F_i} = -\partial E_\theta / \partial \mathbf{r_i}$ | Supervised learning of physical quantities such as energy and forces | Input: element types, atomic coordinates and boundary conditions.<br>Output: energy and atomic forces; may include auxiliary physical quantities such as stress. | Machine-learning approximation of first-principles potential energy surfaces for structural relaxation and molecular dynamics. | Energy reference, units, stress convention, applicable structural domain and simulation interface. | Energy invariance and force equivariance; extensivity; modeling of local or long-range interactions; energy–force consistency. | CHGNet; MatterSim; SphereNet | MPtrj; MD17 | Separate MAE/RMSE for energy, forces and stress; simulation stability and errors in related physical quantities. |
| Electronic structure | Atomic structure and electronic-state specification → electronic-structure representation;<br>$\rho(\mathbf{r}) = f_\theta(X, \mathbf{r})$ | Supervised prediction; variational learning, etc. | Input: atomic structure and electronic-state specification; query coordinates or basis-set information added according to the representation.<br>Output: electron density, Hamiltonian, wavefunction, energy spectrum, etc. | Surrogate modeling of electronic structure and acceleration of electronic-structure calculations. | Electronic state, reference method, basis set or grid, boundary conditions and representation conventions. | Geometric symmetries, together with physical constraints appropriate to the predicted quantity. | InfGCN | QM9_ES; MD17_ES; MP_ES; OMol25_MC_ES | Evaluated according to the predicted quantity: density-field error, error in the integrated electron count, matrix error, spectrum or energy error. |
| Spectrum elucidation | Observed signal → candidate structures, structural features or rankings;<br>$X \sim p_\theta(X \mid S)$ | Conditional generation, retrieval or candidate ranking | Input: unimodal or multimodal spectroscopic data and auxiliary constraints.<br>Output: candidate structures, structural features or rankings. | Inference of molecular or materials structures from spectroscopic information. | Spectroscopic modality, peak or line representation, auxiliary constraints, structural-equivalence criteria and multiplicity of solutions. | Spectrum–structure correlation; chemical validity; modality-specific signal patterns. | DiffNMR | MSD_NMR | Top-k structure match rate, structural similarity; spectral consistency evaluated where necessary. |

| Task family | Computational abstraction | Learning and modeling paradigm | Representations and inputs/outputs | Concrete materials-science instances | Task specificity | Main inductive biases and physical constraints | Representative models | Representative datasets | Evaluation metrics |
|---|---|---|---|---|---|---|---|---|---|
| Spectrum enhancement | Degraded observation → enhanced observation; $Y_{degraded} \rightarrow Y_{enhanced}$ | Supervised, self-supervised or unsupervised restoration | Input: degraded spectral signals or microscopy images.<br>Output: denoised, reconstructed or resolution-enhanced observations. | Improving the quality of experimental characterization to support subsequent structural analysis. | Imaging or spectroscopic modality, degradation mechanism, noise distribution and definition of the reference signal. | Observation physics, noise statistics, spatial or frequency-domain correlations. | SFIN | SFIN-HAADF/BF | Images: PSNR, SSIM.<br>Spectral signals: reconstruction error, peak-position and peak-intensity errors; downstream-task metrics may be added. |

Notes: 1. The computational abstraction, learning paradigm and inputs/outputs describe the general scope of each task family; the models and datasets list representative implementations in PaddleMaterials and do not imply that the platform supports all methods and outputs of a task family. Task families can overlap and do not form a strictly mutually exclusive classification. 2. Inductive biases and physical constraints are adopted according to the object and the model and do not imply that all models satisfy them explicitly. In MLIPs, energy–force consistency applies to models that obtain forces by differentiating the energy; models that predict forces directly require a separate test of their conservativeness. 3. Metrics are chosen according to the task, with units, normalization, data splits and evaluation protocols stated. Structure-matching metrics must specify the matching rule and the number of samples; where stability is involved, the relaxation method, reference energies and decision thresholds must also be specified. Training losses are not listed together with final task metrics.

## Public benchmarks and execution performance

Supplementary Tables S2–S7 list the complete benchmark values for each panel of Fig. 3 in the main text, and Supplementary Table S8 lists the execution times and speedups for Fig. 4 in the main text.

Supplementary Table S2 | Property prediction benchmark [19–23].

| | **MP2018 (crystal materials) · MAE ↓** | | | | **QM9 (organic molecules) · MAE ↓** | | | | | | | | | | | |
|---|---|---|---|---|---|---|---|---|---|---|---|---|---|---|---|---|
| | **Formation energy (meV atom$^{-1}$)** | **Band gap (eV)** | **Bulk modulus K ($\log_{10}$ (GPa))** | **Shear modulus G ($\log_{10}$ (GPa))** | **Dipole moment μ (D)** | **Polariz-ability α (Bohr$^3$)** | **HOMO energy $\varepsilon$HOMO (meV)** | **LUMO energy $\varepsilon$LUMO (meV)** | **Gap Δε* (meV)** | **Electronic spatial extent ⟨$R^2$⟩ (Bohr$^2$)** | **Zero-point vibrational energy ZPVE (meV)** | **Internal energy $U_0$ (meV)** | **Internal energy U (meV)** | **Enthalpy H (meV)** | **Gibbs free energy G (meV)** | **Heat capacity at constant volume $C_v$ (cal mol$^{-1}$ K$^{-1}$)** |
| MEGNet$^{tf}$ | 28.0 | 0.33 | 0.05 | 0.079 | — | — | — | — | — | — | — | — | — | — | — | — |
| MEGNet$^{paddle}$ | 26.5 | 0.2934 | 0.0585 | 0.0962 | — | — | — | — | — | — | — | — | — | — | — | — |
| iComFormer$^{torch}$ | 18.26 | 0.193 | 0.038 | 0.0637 | — | — | — | — | — | — | — | — | — | — | — | — |
| iComFormer$^{paddle}$ | 18.1 | 0.209 | 0.0416 | 0.0651 | — | — | — | — | — | — | — | — | — | — | — | — |
| DimeNet++$^{tf}$ | 49.0‡ | 0.392‡ | 0.041‡ | 0.068‡ | 0.0297 | 0.0435 | 24.6 | 19.5 | 32.6 | 0.331 | 1.21 | 6.32 | 6.28 | 6.53 | 7.56 | 0.0230 |
| DimeNet++$^{paddle}$ | 23.361 | 0.245299 | 0.03448 | 0.0630 | 0.027170 | 0.039659 | 31.777 | 22.304 | 49.751 | 0.256185 | 1.534 | 7.011 | 7.273 | 7.126 | 7.898 | 0.023507 |
| SphereNet$^{torch}$ | — | — | — | — | 0.0245 | 0.0449 | 22.8 | 18.9 | 31.1 | 0.268 | 1.12 | 6.26 | 6.36 | 6.33 | 7.78 | 0.0215 |
| SphereNet$^{paddle}$ | — | — | — | — | 0.024501 | 0.045156 | 24.492 | 18.843 | 31.758 | 0.260937 | 1.262 | 6.218 | 6.443 | 6.599 | 7.639 | 0.025012 |

Notes: tf, torch and paddle denote the TensorFlow/Keras, PyTorch and PaddlePaddle implementations, respectively; the QM9 reference values in the DimeNet++$^{tf}$ row are taken from Table 2 of the DimeNet++ paper [21] (the authors' TensorFlow implementation); the MP2018 reference values marked ‡ are taken from [23]. * The SphereNet gap is defined as LUMO−HOMO. "—" denotes not reported or not applicable. DimeNet++$^{paddle}$ is trained on the moduli in GPa, and its $\log_{10}$ MAEs are computed from its predictions; for K, one of the 341 test structures, whose prediction is negative, is excluded.

Supplementary Table S3 | MLIP benchmark [11, 22, 24].

**a**

| Model | MD17 (organic molecules) · MAE ↓ E: energy (kcal mol⁻¹); F: force (kcal mol⁻¹ Å⁻¹) | | | | | | | | | | | | | | | |
|---|---|---|---|---|---|---|---|---|---|---|---|---|---|---|---|---|
| | Aspirin | | Benzene | | Ethanol | | Malonaldehyde | | Naphthalene | | Salicylic acid | | Toluene | | Uracil | |
| | E | F† | E | F† | E | F† | E | F† | E | F† | E | F† | E | F† | E | F† |
| $\text{SphereNet}^{\text{torch}}$ | 0.364 | 0.43 | 0.590 | 0.178 | 0.294 | 0.208 | 0.104 | 0.34 | 0.680 | 0.178 | 0.298 | 0.36 | 0.164 | 0.155 | 0.700 | 0.267 |
| $\text{SphereNet}^{\text{paddle}}$ | 0.348 | 0.346 | 0.566 | 0.201 | 0.085 | 0.237 | 0.546 | 0.315 | 0.511 | 0.113 | 0.139 | 0.294 | 0.232 | 0.124 | 0.332 | 0.255 |

**b**

| Model | MPtrj 2022.9 (crystal materials) · error metrics ↓ | | | |
|---|---|---|---|---|
| | Energy MAE (meV atom⁻¹) | Force component MAE (meV Å⁻¹) | Stress component MAE (GPa) | Magnetic moment MAE (μB) |
| $\text{CHGNet}^{\text{torch}}$ | 30 | 77 | 0.348 | 0.032 |
| $\text{CHGNet}^{\text{paddle}}$ | 26.7203 | 60.9013 | 0.27484 | 0.03923 |

**c**

| Model | MPtrj-random-1k (crystal materials) · error metrics ↓ | | |
|---|---|---|---|
| | Energy MAE (eV atom⁻¹) | Force vector L2 error (eV Å⁻¹) | Stress error‡ (GPa) |
| $\text{MatterSim}^{\text{torch}}$ | 0.0304 / 0.0236 | 0.1493 / 0.1094 | 1.6843 / 1.3466 |
| $\text{MatterSim}^{\text{paddle}}$ | 0.0304 / 0.0236 | 0.1493 / 0.1094 | 1.6843 / 1.3466 |

Notes: torch and paddle denote the PyTorch and PaddlePaddle implementations, respectively; "-" denotes not reported. The $\text{CHGNet}^{\text{paddle}}$ force and stress MAEs are averaged over Cartesian components and over the nine tensor components, respectively, and the magnetic-moment MAE counts only labeled atoms. ‡ MatterSim results are listed in the order 1M/5M. The $\text{MatterSim}^{\text{torch}}$ row was re-evaluated with the original implementation under the same protocol, using the same released weights and the same 1,000 structures as $\text{MatterSim}^{\text{paddle}}$; the per-structure stress differences between the two implementations are below $10^{-4}$ GPa. The energy error is the per-atom absolute error, the force error is the within-structure mean of the L2 errors of the atomic force vectors, and the stress error is the Frobenius norm of the difference tensor; all three are then averaged over structures. † SphereNet reference force MAEs are the values reported for the force-loss weight p = 100, with the training loss $L = L_E + pL_F$, as for Paddle. Force MAEs are averaged over atomic Cartesian components. The energy MAEs of $\text{SphereNet}^{\text{torch}}$ were obtained by retraining the original code with the same protocol as $\text{SphereNet}^{\text{paddle}}$ (p = 100); because the original code overflows on a few configurations, frames with non-finite outputs or with energy or force errors above 100 kcal mol⁻¹ (Å⁻¹) were excluded: 0, 1, 16, 73, 5, 14, 13 and 2 frames for aspirin, benzene, ethanol, malonaldehyde, naphthalene, salicylic acid, toluene and uracil, respectively.

Supplementary Table S4 | Structure generation benchmark [2, 25].

| | MP20 (crystal materials) · composition-conditioned structure prediction | | MP20 (crystal materials) · de novo generation | | | | |
|---|---|---|---|---|---|---|---|
| | **Match rate ↑ (%)** | **Normalized RMSD ↓ (dimensionless)** | **S.U.N. ↑ (%)** | **Stable ↑ (%)** | **Unique ↑ (%)** | **Novel ↑ (%)** | **Relaxation RMSD ↓ (Å)** |
| DiffCSP$^{torch}$ | 51.49 | 0.0631 | 18.0 | 43.0 | 99.7 | 72.3 | 0.364 |
| DiffCSP$^{paddle}$ | 51.72 | 0.0591 | 13.7 | 37.2 | 99.2 | 73.6 | 0.429 |
| MatterGen$^{torch}$ | 38.66 | 0.0586 | 21.9 | 44.3 | 99.4 | 74.4 | 0.180 |
| MatterGen$^{paddle}$ | 36.05 | 0.0597 | 23.8 | 48.4 | 99.6 | 72.9 | 0.188 |

Notes: torch and paddle denote the PyTorch and PaddlePaddle implementations, respectively. For composition-conditioned structure prediction, both MatterGen implementations are trained for 900 epochs with the official CSP recipe (the original implementation is the official code, retrained), and each test structure is sampled once. For de novo generation, each implementation generates 1,000 structures, which are relaxed and evaluated with the same MatterSim-assisted workflow (5M weights); the original implementations use the officially released weights (mp_gen for DiffCSP and mp_20_base for MatterGen). The DiffCSP percentages use the evaluable structures as the denominator (927 for the original implementation and 890 for PaddleMaterials): the energy above the convex hull cannot be computed for structures containing an element that has no elemental entry in the reference dataset, so these structures are not counted (73 and 110, respectively); with all 1,000 structures as the denominator, S.U.N. is 16.7% and 12.2%, and Stable is 39.9% and 33.1%, respectively. When the PaddleMaterials sampler loads the official DiffCSP weights, S.U.N. and Stable are 18.1% and 45.1%.

Supplementary Table S5 | Electron-density prediction benchmark [26].

| **Model** | Electron-density NMAE ↓ (%) | | | | | | | |
|---|---|---|---|---|---|---|---|---|
| | **QM9 (organic molecules)†** | **MP (crystal materials)** | **MD17 (organic molecules)‡** | | | | | |
| | | | **Benzene** | **Ethane** | **Ethanol** | **Malonaldehyde** | **Phenol** | **Resorcinol** |
| InfGCN$^{torch}$ | 0.93 | 8.98 | 9.5312 | 13.9345 | 12.2454 | 15.0456 | 8.9970 | 8.1682 |
| InfGCN$^{paddle}$ | 0.9315 | 7.7784 | 3.5418 | 4.0910 | 4.4674 | 6.2800 | 3.5976 | 3.4942 |

Notes: torch and paddle denote the PyTorch and PaddlePaddle implementations, respectively; NMAE is given in %. † Both QM9 rows use the unrotated protocol, that is, no random rotation is applied to the molecules and density fields at test time, and are evaluated on all grid points of the last 1,600 test molecules. ‡ Neither MD17 row applies rotation; both are evaluated on all $50^3$ grid points and use the publicly released test split of each molecule (500 configurations for benzene, 201 for ethane, 1,000 for ethanol, 400 for malonaldehyde, 1,200 for phenol and 1,000 for resorcinol); InfGCN$^{torch}$ gives the results of the original code retrained under the same protocol (random seed 42, the original md.yml configuration, 2,000 training steps).

Supplementary Table S6 | Spectrum enhancement benchmark [27].

| | SFIN synthetic HAADF (crystal materials) | | | | SFIN synthetic BF (crystal materials) | | | |
|---|---|---|---|---|---|---|---|---|
| | **Enhancement PSNR ↑ (dB)** | **Enhancement SSIM ↑ (dimensionless)** | **Detection PSNR ↑ (dB)** | **Detection SSIM ↑ (dimensionless)** | **Enhancement PSNR ↑ (dB)** | **Enhancement SSIM ↑ (dimensionless)** | **Detection PSNR ↑ (dB)** | **Detection SSIM ↑ (dimensionless)** |
| SFIN$^{torch}$ | 38.48 | 0.9582 | 28.67 | 0.9642 | 33.24 | 0.9885 | 25.77 | 0.9430 |
| SFIN$^{paddle}$ | 38.7396 | 0.962186 | 28.7011 | 0.9638 | 32.3710 | 0.9788 | 25.7358 | 0.9428 |

Notes: torch and paddle denote the PyTorch and PaddlePaddle implementations, respectively. The reference enhancement and detection results are taken from ref. [27].

Supplementary Table S7 | Spectrum elucidation benchmark [28].

| MSD · HAC ≤ 15 (organic molecules) · exact reconstruction accuracy ↑ (%) | | | | | | |
|---|---|---|---|---|---|---|
| **Model** | **Known formula Top-1** | **Known formula Top-5** | **Known formula Top-10** | **Unknown formula Top-1** | **Unknown formula Top-5** | **Unknown formula Top-10** |
| DiffNMR$^{paddle}$ | 68.26 | 75.04 | 80.27 | 58.83 | 66.61 | 71.39 |

Notes: paddle denotes the PaddlePaddle implementation. The results are taken from Table 1 of ref. [28] and give the exact reconstruction accuracy (%) for MSD molecules with a heavy-atom count HAC ≤ 15; the input is the combined $^{1}H$ and $^{13}C$ spectra, with similarity filtering and retrieval-free initialization, and Top-1, Top-5 and Top-10 results are reported separately for the known- and unknown-formula settings.

Supplementary Table S8 | CINN execution performance of PaddleMaterials and eager comparison with the original implementations.

| Model group | Eager performance comparison | | | CINN compiled execution | | | | |
|---|---|---|---|---|---|---|---|---|
| | Original (ms) | Paddle (ms) | Speedup [range] | n | First call (s) | Eager warm (ms) | CINN warm (ms) | Speedup[range] |
| CHGNet$^{paddle}$ | 35.87 | 26.01 | 38% [30%, 47%] | 1 | 143.703 | 39.511 | 19.048 | 107% [107%, 107%] |
| ComFormer$^{paddle}$ | 5.11 | 8.64 | −38% [−46%, −24%] | 8 | 54.159 | 8.832 | 10.408 | −17% [−37%, −6%] |
| DiffCSP$^{paddle}$ | 30.57 | 37.67 | −18% [−27%, −5%] | 1 | 22.063 | 42.967 | 32.509 | 32% [32%, 32%] |
| DiffNMR$^{paddle}$ | — | 90.03 | — | 1 | 132.210 | 19,624.969 | 17,736.939 | 11% [11%, 11%] |
| DimeNet++$^{paddle}$ | 18.75 | 21.07 | −16% [−35%, −4%] | 4 | 186.711 | 27.770 | 11.799 | 126% [98%, 176%] |
| InfGCN$^{paddle}$ | 65.26 | 363.54 | −82% [−84%, −81%] | — | — | — | — | — |
| MatterGen$^{paddle}$ | 141.02 | 171.21 | −16% [−19%, −11%] | 16 | 266.045 | 368.771 | 512.713 | −29% [−35%, −11%] |
| MatterSim$^{paddle}$ | 41.63 | 33.95 | 23% [4%, 39%] | 2 | 163.861 | 39.482 | 14.154 | 185% [153%, 216%] |
| MEGNet$^{paddle}$ | 223.52 | 11.73 | 1751% [1592%, 1858%] | 8 | 49.942 | 12.442 | 20.326 | −37% [−49%, −25%] |
| SFIN$^{paddle}$ | 18.98 | 37.89 | −50% [−53%, −46%] | 4 | 23.689 | 38.419 | 17.536 | 123% [76%, 166%] |
| SphereNet-MD17$^{paddle}$ | 53.59 | 43.56 | 20% [14%, 26%] | 8 | 241.368 | 56.755 | 17.303 | 224% [124%, 336%] |
| SphereNet-QM9$^{paddle}$ | 14.23 | 20.43 | −28% [−32%, −20%] | 12 | 135.946 | 21.272 | 14.755 | 47% [29%, 67%] |

Notes: The eager-comparison columns were measured separately: both sides use the same inputs; times are medians of warm single calls, and the speedup is the original time divided by the Paddle time minus 1, expressed as a percentage, so positive values mean that the Paddle implementation is faster. For MatterSim, the 1M and 5M model instances were measured; times are group medians, the speedup is computed per model instance and its median is taken, and the square brackets give the minimum and maximum speedups over all measurement blocks. Inputs: the 8-atom $LiMnO_2$ cell for CHGNet, ComFormer, DimeNet++, MatterSim and MEGNet; generation of an 8-atom structure with 2 reverse diffusion steps for DiffCSP and MatterGen, generation of a molecular graph from the NMR conditions of the toluene case with 2 reverse diffusion steps for DiffNMR; the aspirin molecule (21 atoms) for SphereNet-MD17; a 17-atom QM9 molecule for SphereNet-QM9; a 15-atom QM9 molecule with 32,768 of its native grid points for InfGCN; and a 256×256 HAADF image for SFIN. These inputs differ from those of the public matrix used in the CINN columns, so the Paddle times in the eager comparison need not equal the eager warm times in the CINN columns. DiffNMR has only a Paddle implementation, and InfGCN is not part of the public CINN matrix; the corresponding cells are marked "—". In the CINN columns, the first CINN call time includes compilation and the first execution; the warm time is the median of 10 calls after the first call. n is the number of model instances tested in each group, 65 in total. Each time column gives the median over the model instances in a group; the speedup is computed for each model instance (the Eager warm time divided by the CINN warm time minus 1, expressed as a percentage, with positive values meaning that CINN is faster) and its median over the group is given, and the square brackets give the minimum and maximum speedups of the model instances in the group.